\documentclass[11pt]{article}

\usepackage[final]{acl}

\usepackage{times}
\usepackage{latexsym}

\usepackage[T1]{fontenc}

\usepackage[utf8]{inputenc}

\usepackage{microtype}

\usepackage{inconsolata}

\usepackage{graphicx}

\usepackage{colortbl}

\newcommand{\lightmodelrule}{\arrayrulecolor{black!20}\midrule\arrayrulecolor{black}}

\usepackage{listings}
\usepackage{booktabs}
\usepackage{multirow}
\usepackage{tikz}
\usetikzlibrary{arrows.meta, shapes.geometric, positioning}

\usepackage{tcolorbox}
\tcbuselibrary{breakable,skins}

\definecolor{PromptBg}{HTML}{F9F9F9}
\definecolor{PromptFrame}{HTML}{CCCCCC}

\definecolor{mdcodebg}{RGB}{246, 248, 250}
\lstdefinestyle{mdcode}{
    backgroundcolor=\color{mdcodebg},
    basicstyle=\ttfamily\footnotesize,
    frame=single,
    framerule=0pt,
    rulecolor=\color{mdcodebg},
    breaklines=true,
    breakatwhitespace=true,
    columns=fullflexible,
    upquote=true,
    showstringspaces=false,
    xleftmargin=1em,
    xrightmargin=1em,
    aboveskip=1em,
    belowskip=1em,
    literate=
      {→}{$\to$}{1}
      {Δ}{$\Delta$}{1}
      {—}{---}{1}
      {–}{--}{1},
    postbreak=\mbox{\textcolor{red}{$\hookrightarrow$}\space},
}

\newtcolorbox{promptbox}[1][]{
    breakable,
    colback=PromptBg,
    colframe=PromptFrame,
    boxrule=0.5pt,
    arc=2pt,
    left=4pt, right=4pt, top=4pt, bottom=4pt,
    #1
}

\newcommand{\symbolimg}[2][0.3cm]{%
  \ensuremath{\vcenter{\hbox{\includegraphics[height=#1]{#2}}}}%
}
\newcommand{\openaiicon}{\symbolimg[0.30cm]{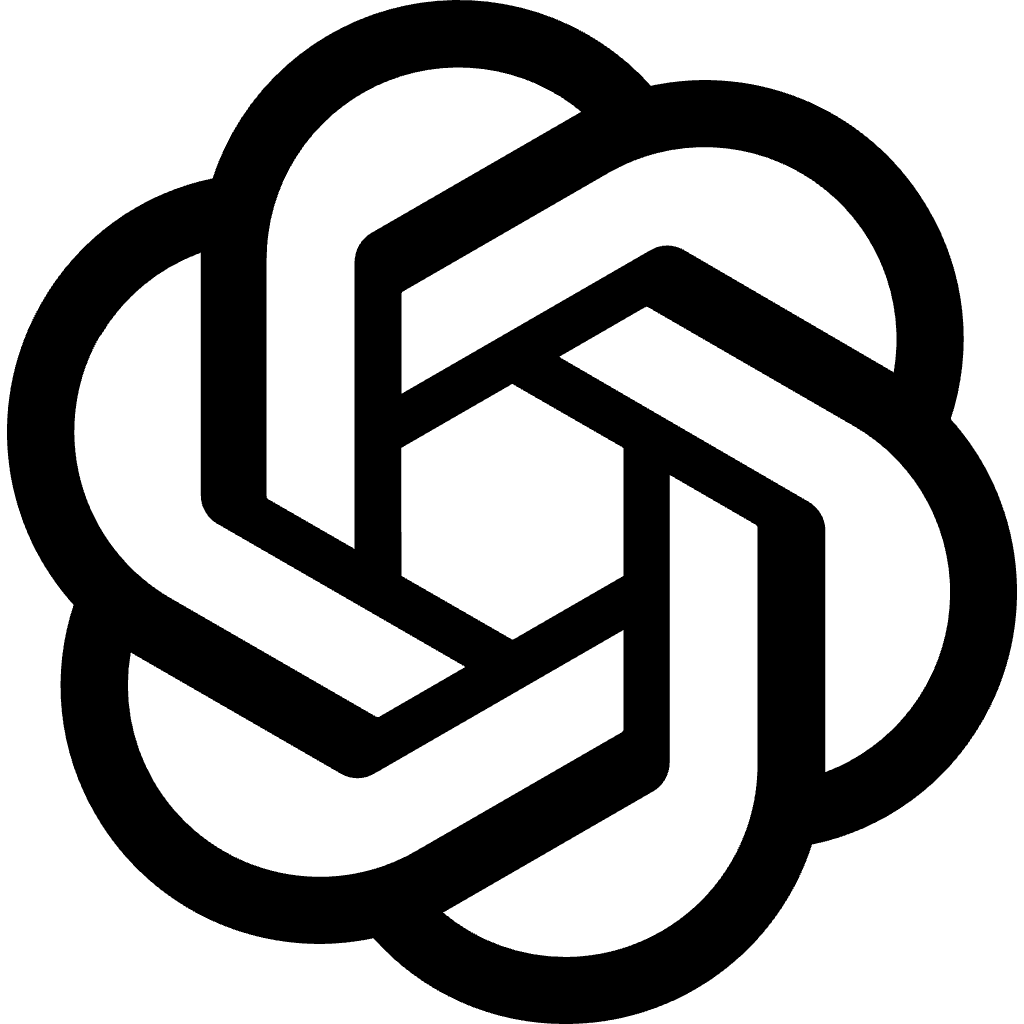}}
\newcommand{\claudeicon}{\symbolimg[0.30cm]{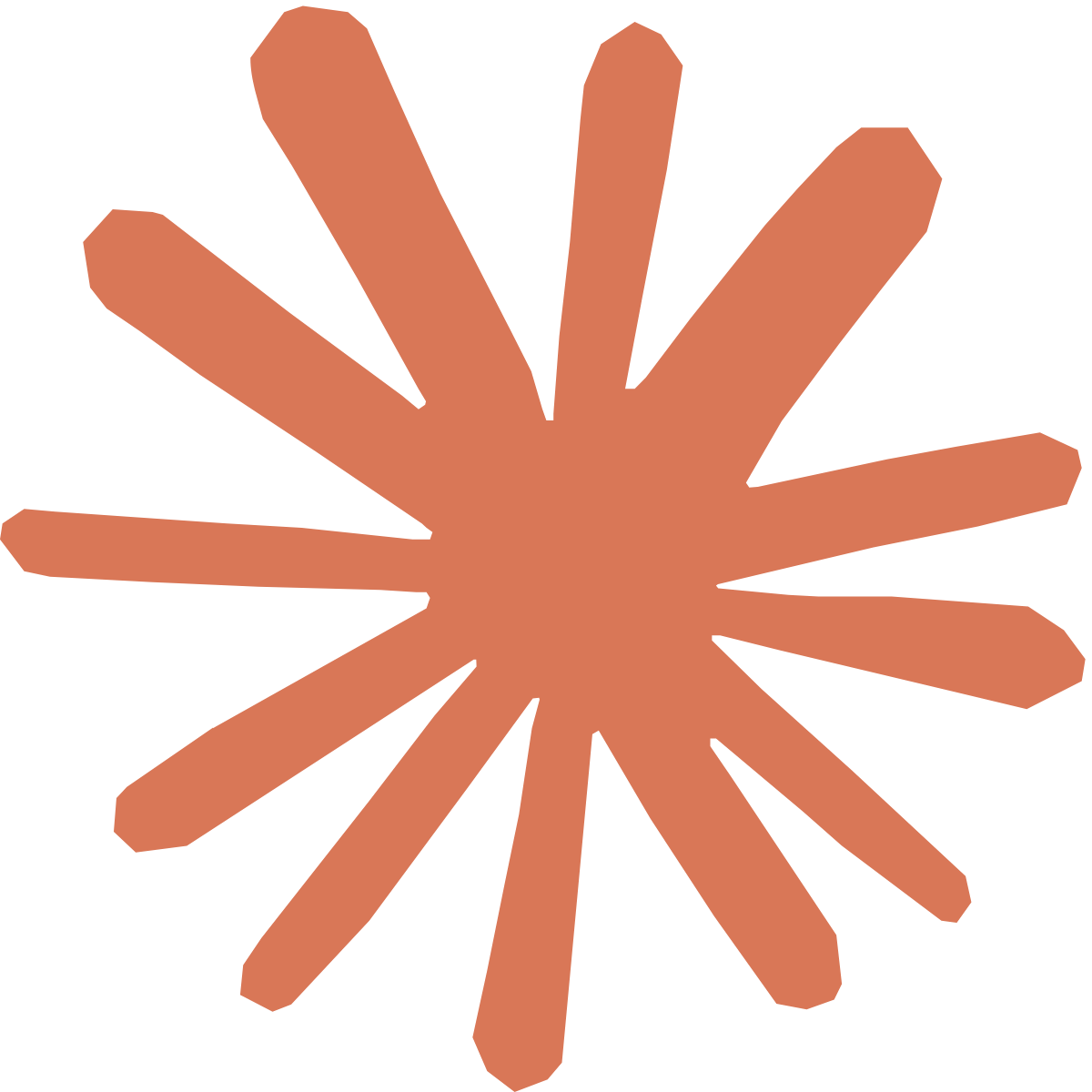}}
\newcommand{\geminiicon}{\symbolimg[0.30cm]{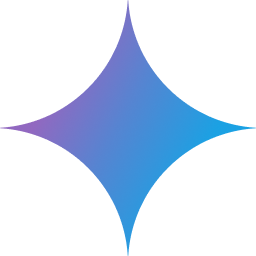}}
\newcommand{\gemmaicon}{\symbolimg[0.30cm]{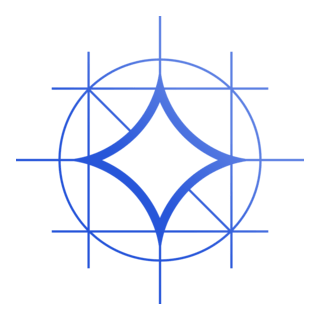}}
\newcommand{\mistralicon}{\symbolimg[0.30cm]{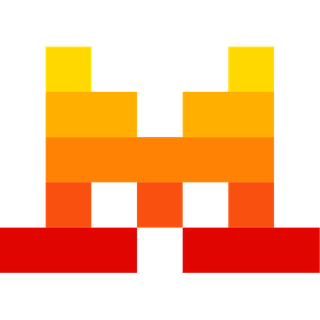}}
\newcommand{\qwenicon}{\symbolimg[0.30cm]{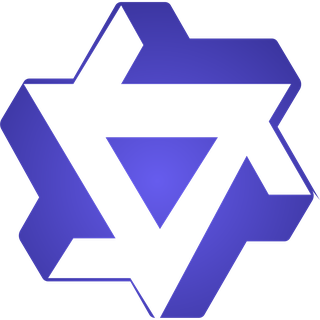}}

\title{Larger Context Window, Fewer Overcorrections: Optimizing Prompts and Batching for Minimal-Edit Grammatical Error Correction\thanks{Accepted for publication at EMNLP 2026 (Findings).}}

\author{
  Kateryna Karpo$^{\upsilon,\sigma}$\thanks{First author.} \quad Artem Chernodub$^{\zeta}$ \\
  $^{\upsilon}$Ukrainian Catholic University \quad $^{\sigma}$YouScan \quad $^{\zeta}$Zendesk \\
  {\normalfont\texttt{kateryna.karpo@ucu.edu.ua} \quad \texttt{a.chernodub@gmail.com}}
}

\begin{document}
\maketitle

\begin{abstract}
Minimal-edit Grammatical Error Correction (GEC) is a challenging task for zero- and few-shot prompted Large
Language Models (LLMs), which systematically overcorrect and degrade $F_{0.5}$ by rewriting well-formed spans.
While fine-tuning provides an effective solution, it imposes substantial infrastructure demands. We introduce a prompt-based approach
that closes the gap to fine-tuned models through three advances in GEC prompting methodology. First, we introduce taxonomy-based
instructions to enforce minimal-edit constraints with a comprehensive list of grammatical error rules,
equipping the LLM with a bounded, metric-aligned scope of correctable edits, which benefits the
strongest models while remaining model-dependent overall. Second, we show that batching
multiple uncorrected sentences into a single input context acts as a targeted regularizer against
overcorrection, systematically reducing the edit rate across diverse LLM families; we hypothesize this arises
from attention dilution effect induced by the bounded capacity of self-attention scores. Finally, LLM-assisted Prompt
Optimization refines these instructions. Powered by Gemini 3.1-Pro, our prompt achieves $F_{0.5}=78.32$ on the BEA-2019
test set — establishing a new prompt-based SOTA while shrinking the gap to the fine-tuned single-model
SOTA \citep{staruch2025adapting} to a mere $0.38$ points. Code, prompts, and outputs are publicly
available.\footnote{This work was conducted as part of Kateryna Karpo's M.Sc. thesis at
the Ukrainian Catholic University, Faculty of Applied
Sciences.}\footnote{\label{fn:repo}\url{https://github.com/katerynkarpo/llm-en-gec}}
\end{abstract}

\section{Introduction}
\label{sec:introduction}

Grammatical Error Correction (GEC) has two evaluation paradigms
\citep{bryant2019bea}. Fluency-oriented correction permits extensive lexical substitution and
stylistic restructuring to produce native-sounding text \citep{napoles2017jfleg}. Minimal-edit correction instead addresses
grammatical, spelling, and punctuation errors while preserving the original
phrasing \citep{leacock-etal-2014-automated}. Such minimal, targeted corrections are essential in educational applications,
where the goal is to guide learners in revising their own errors while preserving intent \citep{nicholls2003cambridge},
and controlling LLMs to produce them remains difficult \citep{vajjala-etal-2026-opportunities}.

Earlier work on minimal-edit GEC relied on supervised fine-tuning of sequence taggers
\citep{awasthi-etal-2019-parallel, omelianchuk2020gector, tarnavskyi2022ensembling} and sequence-to-sequence models
\citep{junczys-dowmunt-etal-2018-approaching,grundkiewicz-etal-2019-neural,kiyono2019empirical,rothe2021simple}. Both approaches face recall bottlenecks: taggers are bound by finite edit
spaces, while sequence-to-sequence cross-attention induces a copying bias. Their conservative span-local bias still
aligns naturally with the precision-weighted $F_{0.5}$ objective and sets a strict behavioral baseline for prompt-based
methods.

\citet{alikaniotis-raheja-2019-unreasonable} demonstrated that early GPT models could perform GEC
without task-specific training.
With the rise of instruction-tuned, decoder-only LLMs \citep{ouyang2022training} such as ChatGPT, zero- and few-shot
prompting became a widely adopted paradigm for text editing without task-specific fine-tuning. Out-of-the-box prompted
LLMs rapidly reached state-of-the-art results on tasks that reward free-form rewriting, such as fluency GEC
\citep{loem2023exploring} and text simplification \citep{vadlamannati-sahin-2023-metric}. Applied to minimal-edit GEC
with vanilla instructions (e.g., \texttt{``Correct the grammatical errors''}), the same models systematically
overcorrect in both zero- and few-shot settings \citep{fang2023chatgpt, coyne2023analyzing}, rewriting well-formed
spans with synonym swaps, stylistic polishing, or clause rephrasings that add no actual value.

Recent prompt-based efforts on minimal-edit GEC explored richer prompting strategies, from
error-pattern-aware in-context example selection \citep{tang-etal-2024-ungrammatical,li-etal-2025-explanation} to
detection-before-correction prompting \citep{li-wang-2024-detection} and automatic prompt optimization
\citep{chernodub2025apio}. Even so, prompt-based systems still lag behind the supervised state of the art, so the
mainstream route to enforcing the minimal-edit constraint remains infrastructure-heavy fine-tuning:
\citet{omelianchuk2024pillars} adapt LLaMA-2 to reduce overcorrection, \citet{liang2025edit} align the pretraining
objective with the minimal-edit principle through edit-wise preference optimization for LLaMA2 and Mistral-v0.1, and
\citet{staruch2025adapting} set the current single-model state of the art on BEA-2019 ($F_{0.5}=78.70$) by fine-tuning
Gemma 2 with a schedule that progressively raises the share of unedited examples, reversing earlier sequence-to-sequence pipelines
that filtered out error-free sentences \citep{junczys-dowmunt-etal-2018-approaching,grundkiewicz-etal-2019-neural}.

 We aim to close the gap between fine-tuned GEC systems and prompt-based, API-accessed LLMs, which
matters for individuals and organizations that cannot afford fine-tuning of open-source LLMs due to engineering or regulatory
constraints. Our hypothesis is that the increased instruction-following capabilities of the latest LLMs are already sufficient to respect the
minimal-edit GEC constraints without the need of weight updates. Recent results support this, but also expose
its limit: with a declarative minimal-edit instructions, Claude Sonnet 4.5 reaches $F_{0.5}=64.91$ on the BEA-2019 test
set \citep{turker-eryigit-2026-instruction}, yet still trails the fine-tuned single-model SOTA by $13.79$ points. We
believe the declarative style is one of the reasons: it states a goal without saying which edits are admissible. Our
taxonomy-based instructions supply that missing specification, bounding the edit space to the 25 metric-aligned
ERRANT categories \citep{bryant-etal-2017-automatic}. Once instructions reach this level of quality, we search for
model-specific instruction refinements via LLM-assisted prompt optimization.
Last but not least, we introduced batching to reduce latency and cost \citep{cheng-etal-2023-batch}, but unexpectedly
yielded consistent quality improvements for minimal-edit GEC, motivating our systematic study of how batch size
affects edit rates and correction quality.
We evaluate our approach on six commercial LLMs and one open-weight model across two minimal-edit benchmarks.

Our primary contributions are:

1. We set a new state of the art for prompt-based minimal-edit GEC. With Gemini 3.1-Pro and our best prompt \ref{prompt:apo:gemini}, we
achieve $F_{0.5}=78.32$ on the BEA-2019 test set, the first time an out-of-the-box prompted LLM has closed the gap to
the best fine-tuned single model \citep{staruch2025adapting} to a mere $0.38$ $F_{0.5}$ points. We also report a
competitive $F_{0.5}=67.08$ on the CoNLL-2014 test set.

2. We introduce taxonomy-based GEC instructions layered on a minimal-edit baseline to provide a bounded,
metric-aligned scope of edits that constrains the strongest models productively and acts as noise for the
others. Defining errors via a concrete list addresses the limitations of both the vague
``grammatical error'' framing of vanilla GEC prompts and declarative ``minimal edit'' instructions that ask for
conservative rewriting without explaining how to do so.

3. We identify batching as a targeted regularizer against overcorrection: packing multiple sentences into a single
context reduces the word-level edit rate and consistently improves $F_{0.5}$ scores. We hypothesize this stems from an
attention dilution effect induced by the bounded capacity of self-attention scores.

4. We apply LLM-assisted Prompt Optimization, implemented as a reusable Claude Code agentic skill\footref{fn:repo},
on top of taxonomy-based instructions and batching. The agent iteratively clusters dev-set errors and greedily adopts
prompt revisions that improve the $F_{0.5}$ score.

\section{Related Work}
\label{sec:related_work}

The tendency of generative LLMs to overcorrect was recognized early, shifting the focus of prompt-based GEC toward
finding the right wording and structure to enforce the minimal-edit constraint. Early evaluations by
\citet{fang2023chatgpt} and \citet{coyne2023analyzing} relied on declarative prompts that asked models to maximize
source preservation. \citet{loem2023exploring} then showed that prompt wording alone steers GPT between minimal-edit
and fluency-edit regimes, a finding \citet{davis2024prompting} extend to open-source LLMs, and out-of-the-box LLM
performance also varies sharply with writer proficiency \citep{zeng2024evaluating} and across languages
\citep{katinskaia2024gpt}, factors that earlier fine-tuned systems handled by adapting the GEC model to the
writer's proficiency level and first language \citep{nadejde-tetreault-2019-personalizing}. More recent efforts to reduce overcorrection move beyond basic instructions:
\citet{li-wang-2024-detection} introduce a two-step detection-correction architecture,
\citet{tang-etal-2024-ungrammatical} propose ungrammatical-syntax-based in-context example selection, and
\citet{goto-etal-2026-edit} aggregate multiple decodings via edit-level majority voting.
\citet{turker-eryigit-2026-instruction} report the strongest zero-shot prompted results so far with
declarative minimal-edit instructions, structurally close to our minimal-edits zero-shot prompt
(\ref{prompt:minimal-edits-zero-shot}); introducing the taxonomy (\ref{prompt:taxonomy-zero-shot}) is what
distinguishes our instructions. This minimal-edits + taxonomy design is adapted from our prior work on Ukrainian
\citep{karpo-chernodub-2026-far}, evaluated on the UNLP 2023 shared task benchmark
\citep{syvokon-romanyshyn-2023-unlp}, built on the UA-GEC corpus \citep{syvokon-etal-2023-ua}.
Rather than relying on multi-step
pipelines or inference-time voting, we constrain the editing scope natively through GEC taxonomy-based instructions.

Prior work treats batching purely as an inference-throughput optimization \citep{cheng-etal-2023-batch}, in GEC as
well \citep{masciolini2025multigec}. In
the broader LLM literature, expanding the input context is typically associated with quality degradation driven by
attention dilution, including ``lost in the middle'' effects and reasoning failures
\citep{qin-etal-2022-devil, liu2023lost, liu2023flipflop}. To our knowledge, no earlier work uses the input batch as a
targeted regularizer to actively reduce overcorrection in minimal-edit GEC.

Automatic Prompt Optimization (APO) has emerged as a popular alternative to manual prompt engineering, where an LLM
iteratively rewrites the instruction it executes \citep{ramnath-etal-2025-systematic}. Within GEC,
\citet{chernodub2025apio} apply this paradigm to GPT-4o, treating the prompt as a learnable list of instructions
inferred and refined automatically. Our procedure is instead a semi-automatic Claude Code skill that rewrites the
entire prompt under user feedback at each iteration.

\section{Proposed Method}
\label{sec:method}
We present a prompt-based minimal-edit GEC framework combining GEC taxonomy-grounded prompts and batched inference,
further refined via LLM-assisted Prompt Optimization.

\paragraph{Prompting strategies.}
\label{sec:prompting-strategies}
We evaluate five manually engineered prompts (\ref{prompt:vanilla-zero-shot}--\ref{prompt:taxonomy-few-shot}), varying
whether few-shot examples and the 25 ERRANT error categories are included. The strongest one, ``minimal-edits few-shot
+ taxonomy'' (\ref{prompt:taxonomy-few-shot}), is then optimized separately for medium-capacity models
(Section~\ref{sec:experimental_setup}), yielding three LLM-optimized prompt variants tailored to specific model
families (OpenAI GPT, Anthropic Claude, Google Gemini), as well as the open-weight Qwen3-8B model.

\begin{enumerate}
\item \textbf{Vanilla zero-shot prompt} (Appendix~\ref{prompt:vanilla-zero-shot}): a generic instruction to fix
grammatical and spelling errors, adapted from \citet{coyne2023analyzing}, with no further constraints and no examples:
\texttt{``Reply with a corrected version of the input sentence with all grammatical and spelling errors fixed. If
there are no errors, reply with a copy of the original sentence.''}

\item \textbf{Minimal-edits zero-shot prompt} (Appendix~\ref{prompt:minimal-edits-zero-shot}): declares the
minimal-edit constraint directly (\texttt{``Make MINIMAL, PRECISE edits to fix errors. DO NOT rewrite or
paraphrase.''}), plus a short rule list banning paraphrasing, fluency improvements, and stylistic changes, aligned
with \citet{loem2023exploring} and \citet{davis2024prompting}. No grammatical error categories are enumerated, so the
model is left to decide what counts as an error.

\item \textbf{Minimal-edits few-shot prompt} (Appendix~\ref{prompt:minimal-edits-few-shot}): keeps the minimal-edit
constraint and rule list of prompt~(\ref{prompt:minimal-edits-zero-shot}) and adds 8 source/target correction pairs
sampled from the BEA-2019 train split, presented as a flat list of demonstrations without taxonomic grouping or
category labels.

\item \textbf{Minimal-edits zero-shot + taxonomy prompt} (Appendix~\ref{prompt:taxonomy-zero-shot}): supplements the
minimal-edit constraint with the 25 ERRANT error categories \citep[Table~2]{bryant-etal-2017-automatic} to provide
explicit taxonomic guidance, addressing the ambiguity left by prompt~(\ref{prompt:minimal-edits-zero-shot}). The 25
categories are grouped into word-level, mechanical, and other classes; each is listed with its ERRANT label, a short
gloss, and an illustrative substitution, e.g.\ \texttt{``6.~DET: Wrong/missing/extra determiner (the$\to$a,
$\emptyset\to$the, the$\to\emptyset$)''}, \texttt{``20. PUNCT: Punctuation errors (!$\to$., missing commas, extra
periods)''}. Any edit outside these 25 categories is by construction off-task, and the prompt declares this
explicitly.

\item \textbf{Minimal-edits few-shot + taxonomy prompt} (Appendix~\ref{prompt:taxonomy-few-shot}): augments
prompt~(\ref{prompt:taxonomy-zero-shot}) with 8 source/target correction pairs sampled from the BEA-2019 train split.

\item \textbf{Minimal-edits few-shot + taxonomy + optimized prompt}
(\ref{prompt:apo:qwen}--\ref{prompt:apo:gemini}): a refinement of
prompt~\ref{prompt:taxonomy-few-shot} produced by an LLM-assisted Prompt Optimization for each model family.
\end{enumerate}

\begin{figure}[t!]
  \centering
  \includegraphics[width=\columnwidth]{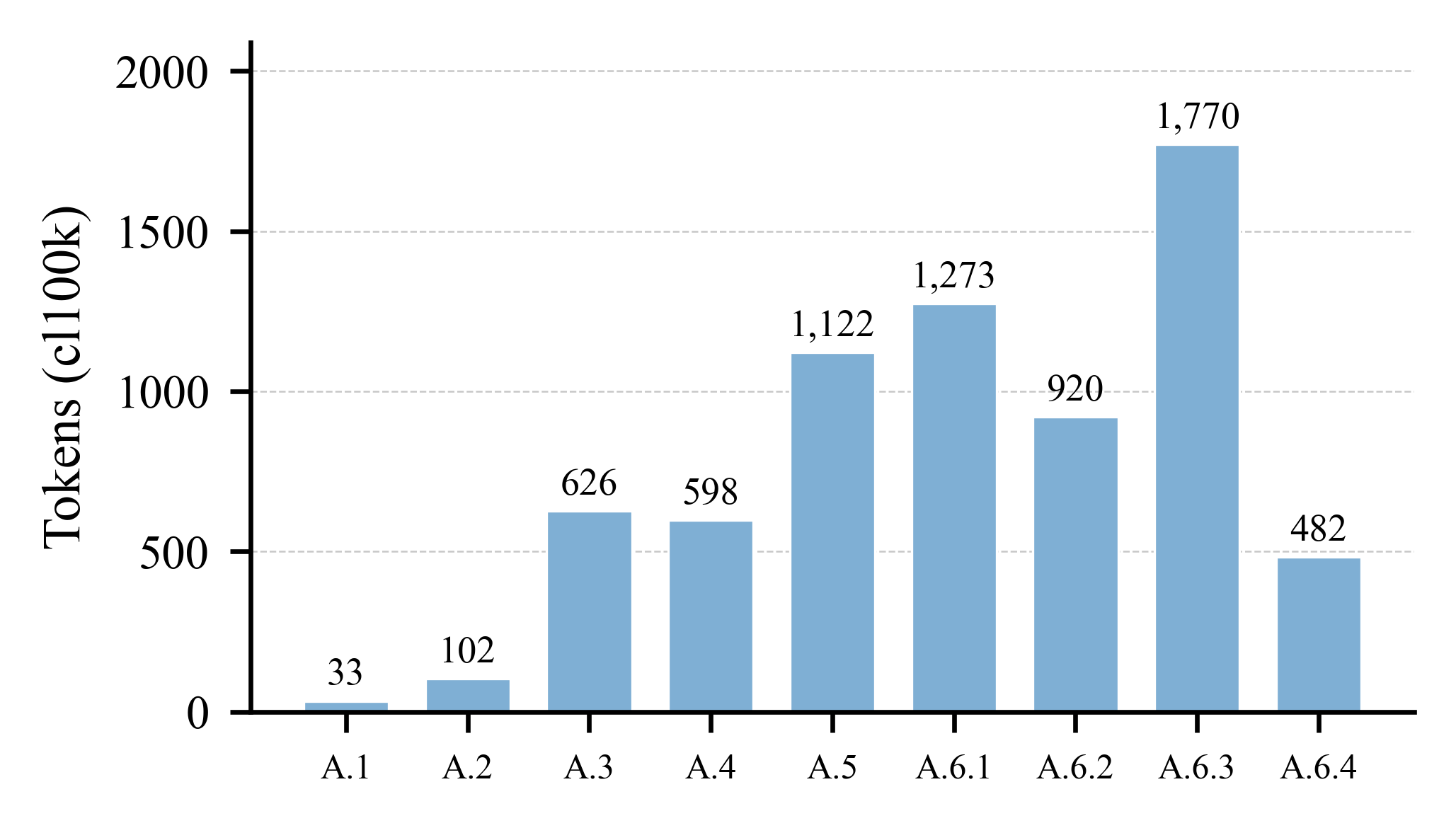}
\caption[Prompt length across prompting strategies]{Prompt length in cl100k\_base
tokens\protect\footnotemark. Manual prompts (\ref{prompt:vanilla-zero-shot}--\ref{prompt:taxonomy-few-shot})
grow progressively longer as constraints and examples are added. In contrast, optimized prompts
(\ref{prompt:apo:qwen}--\ref{prompt:apo:gemini}) yields mixed lengths, either significantly
compressing or further expanding the longest manual baseline.}
  \label{fig:prompt-tokens}
\end{figure}
\footnotetext{\url{https://github.com/openai/tiktoken}}

\paragraph{Batching.}
\label{sec:batching}

We use the same prompts in two modes. The single-sentence mode sets \texttt{<input\_text>} to one source sentence and
processes the input one sentence at a time (we refer these cases as $B=1$). The batched mode, which we advocate as the
default for deploying minimal-edit GEC with prompted LLMs, sets \texttt{<input\_text>} to a numbered, newline-separated
list of $B$ sentences $s_1, \dots, s_B$ drawn from the input stream; the model returns the numbered list of corrected
sentences, and we split it back to the source one-to-one before metric-based evaluation. Inference pipeline
details are in Appendix~\ref{app:inference-pipeline}.

\paragraph{LLM-assisted Prompt Optimization.}
\label{sec:apo}

To refine the prompts, we introduce a human-in-the-loop prompt optimization process
using Claude Opus~4.6. Unlike fully autonomous APO frameworks
\cite{ramnath-etal-2025-systematic, agrawal2025gepareflectivepromptevolution}, our agent periodically requires
researcher guidance, so we classify the approach as LLM-assisted rather than fully automatic.

Starting from Prompt~\ref{prompt:taxonomy-few-shot}, the agent iteratively refines instructions on the BEA-2019 dev
split using a rule-driven greedy search. Each step isolates variables to ensure attributable performance deltas:
\begin{enumerate}
    \item \textbf{Diagnostic Critique:} The agent uses ERRANT to isolate metrics by linguistic category and inspects
    sentence-level errors to generate a targeted natural language diagnosis, mirroring error-driven textual gradients
    and multi-aspect critiques in recent APO literature \cite{protegi, crispo, textgrad}.
    \item \textbf{Targeted Proposal:} The agent proposes exactly one change per iteration, either a specific negative
    constraint (a ``do not change'' rule) or a targeted exemplar. Unlike methods relying on broad search trees or
    complex mutations \cite{promptagent, ramnath-etal-2025-systematic}, we enforce high-precision, human-readable
    heuristic additions targeting specific GEC categories based on ERRANT definitions.
    \item \textbf{Greedy Validation:} The edit is adopted only if it strictly improves the overall dev $F_{0.5}$ score
    and the targeted GEC category moves in the predicted direction.
\end{enumerate}

The loop runs until gains plateau. The optimizer Claude Skill's full code is in Appendix~\ref{app:apo-skill}.

\textbf{Qualitative analysis of optimized prompts.}
Compared with the shared starting point \ref{prompt:taxonomy-few-shot}, the
optimizer reshaped the instruction in a markedly model-dependent way. For GPT-4.1-mini (\ref{prompt:apo:gpt}) it
discarded the taxonomy in favour of two narrowly scoped rules on comma splices and
tense consistency in past narratives, each qualified by explicit exclusions and a leave-it-alone default, keeping
the demonstrations and compressing the prompt from $1{,}122$ to $920$ tokens. For Claude Sonnet~4.6 (\ref{prompt:apo:claude}) it kept the 25 categories
and the demonstrations intact and added two punctuation rule-blocks that enumerate their trigger lexicons
(subordinators and connectors for sentence-initial commas, pronouns and demonstratives for comma splices)
plus hard conditions under which the rule must not fire. For Gemini~3-Flash
(\ref{prompt:apo:gemini}) it replaced the taxonomy with a plain-language list of what to fix and what to leave alone
that prunes the open-ended lexical-choice categories, restricted punctuation to a short whitelist, and supplied five
rule-scoped examples, one deliberately requiring no edit.
For the open-weight Qwen3-8B (\ref{prompt:apo:qwen}) it was far more conservative: the taxonomy and the
eight demonstrations both survive, wording changes are limited to a no-markup output rule and an introductory-comma
rule, and the largest gain came from delivery, re-packaging the demonstrations as a user/assistant chat exchange.

Despite this divergence, the three proprietary-model runs
converged on the same policy: precision-first constraints aimed at the categories with the largest residual dev-set
error, generic demonstrations kept except for Gemini's rule-scoped replacements, the taxonomy treated as optional rather than
essential, and most recall-seeking proposals rejected at validation.
The Qwen3-8B run shares only the precision-first focus: the taxonomy and demonstrations
still help this smaller open-weight model, so both survive untouched. Figure~\ref{fig:prompt-tokens} shows
how prompt length grows across all our nine prompts (five manual and four LLM-optimized variants).

\section{Experimental setup}
\label{sec:experimental_setup}

\paragraph{Datasets and metrics.} For development, we use the restricted track of the BEA-2019 Shared Task
\citep{bryant2019bea}, comprising the Write~\&~Improve and LOCNESS corpora: the official train split (34,308
sentences) for prompt engineering and few-shot sampling, and the dev split (4,384 sentences) for batching
investigation and ablation studies. Final evaluations are performed on the CoNLL-2014 test set \citep{ng2014conll}
(1,312 sentences) using the $M^2$ scorer \citep{dahlmeier2012better} and on the BEA-2019 test set (4,477 sentences) via
CodaBench\footnote{\url{https://www.codabench.org/competitions/10960/}} using ERRANT v3.0.0\footnote{The platform
migrated to ERRANT v3.0.0 due to deprecated dependencies; score differences from legacy v2.0.0 are negligible.}.
Both benchmarks use precision-weighted $F_{0.5}$ as the main metric.
Local evaluations are computed with the \texttt{gec-metrics}
library \citep{goto-etal-2025-gec}.\footnote{\url{https://github.com/gotutiyan/gec-metrics}}

\paragraph{Models.} We focus on commercial LLMs to bring prompt-based GEC to parity with fine-tuned models, serving users who
cannot host open-weight alternatives for engineering or regulatory reasons; open-weight models can be fine-tuned
directly, so prompt optimization matters less for them. We also report the open-weight Qwen3-8B
\citep{yang2025qwen3} as a reproducible baseline.

We evaluate six API-accessed commercial LLMs from three providers (OpenAI, Anthropic, Google), grouped into capability tiers to isolate the effect of model tier from that of
prompting and batching. The
\textit{high-capacity tier} contains the strongest current offering of each provider: GPT-5.4
(\texttt{gpt-5.4-2026-03-05}), Claude Opus~4.6 (\texttt{claude-opus-4.6}), and Gemini~3.1-Pro
(\texttt{gemini-3.1-pro-preview}). The \textit{medium-capacity tier} contains a faster, lower-cost counterpart:
GPT-4.1-mini (\texttt{gpt-4.1-mini-2025-04-14}), Claude Sonnet~4.6 (\texttt{claude-sonnet-4.6}), and Gemini~3-Flash
(\texttt{gemini-3-flash-preview}). \footnote{We report exact snapshot identifiers wherever providers expose them;
Anthropic and Google currently do not, so we record the latest preview alias used at submission time.} \footnote{Inference-time
parameters differ across providers. We set temperature~0 where available, default reasoning effort for Claude,
\texttt{medium} effort for GPT-5.4, and \texttt{thinkingLevel: low} for all Gemini models.}

\paragraph{Inference pipeline.} All seven models are accessed through a single LiteLLM router using Structured Outputs:
every call returns a JSON object validated against a strict Pydantic schema, which removes free-form post-processing
and makes the sentence-to-prediction mapping bijective. The schema, the provider-agnostic \texttt{response\_format}
envelope, and the recovery path for providers that reject strict JSON schema are presented in
Appendix~\ref{app:inference-pipeline}.

\section{Experiments}

\subsection{Prompting strategies}
\label{sec:prompting-strategies-exp}

\begin{table*}[t]
\centering
\footnotesize
\setlength{\tabcolsep}{2pt}
\begin{tabular}{l rrr @{\hskip 6pt} rrr @{\hskip 6pt} rrr @{\hskip 6pt} rrr @{\hskip 6pt} rrr}
\toprule
 & \multicolumn{3}{c}{Vanilla}
 & \multicolumn{3}{c}{Minimal-edits}
 & \multicolumn{3}{c}{Minimal-edits ZS}
 & \multicolumn{3}{c}{Minimal-edits}
 & \multicolumn{3}{c}{Minimal-edits FS} \\
 & \multicolumn{3}{c}{zero-shot (\ref{prompt:vanilla-zero-shot})}
 & \multicolumn{3}{c}{zero-shot (\ref{prompt:minimal-edits-zero-shot})}
 & \multicolumn{3}{c}{+ taxonomy (\ref{prompt:taxonomy-zero-shot})}
 & \multicolumn{3}{c}{few-shot (\ref{prompt:minimal-edits-few-shot})}
 & \multicolumn{3}{c}{+ taxonomy (\ref{prompt:taxonomy-few-shot})} \\
\cmidrule(lr){2-4}\cmidrule(lr){5-7}\cmidrule(lr){8-10}\cmidrule(lr){11-13}\cmidrule(lr){14-16}
\textbf{Model}
 & \textbf{Prec.} & \textbf{Rec.} & $\mathbf{F_{0.5}}$
 & \textbf{Prec.} & \textbf{Rec.} & $\mathbf{F_{0.5}}$
 & \textbf{Prec.} & \textbf{Rec.} & $\mathbf{F_{0.5}}$
 & \textbf{Prec.} & \textbf{Rec.} & $\mathbf{F_{0.5}}$
 & \textbf{Prec.} & \textbf{Rec.} & $\mathbf{F_{0.5}}$ \\
\midrule

\qwenicon~Qwen3-8B
 & 45.30 & \textbf{46.91} & 45.61
 & 56.42 & 27.58 & 46.66
 & 52.56 & 27.49 & 44.45
 & \textbf{58.11} & 30.09 & \textbf{48.98}
 & 57.84 & 29.66 & 48.61 \\

\midrule
\openaiicon~GPT-4.1-mini
 & 44.51 & \textbf{56.57} & 46.49
 & 57.66 & 40.66 & 53.21
 & 54.25 & 40.25 & 50.72
 & \textbf{57.98} & 44.25 & \textbf{54.59}
 & 55.08 & 44.21 & 52.50 \\
\claudeicon~Claude Sonnet 4.6
 & 49.62 & \textbf{53.82} & 50.41
 & \textbf{65.61} & 31.92 & 54.17
 & 63.10 & 34.04 & 53.90
 & 63.13 & 37.43 & \textbf{55.51}
 & 62.66 & 36.48 & 54.79 \\
\geminiicon~Gemini 3-Flash
 & 42.70 & \textbf{57.76} & 45.05
 & 52.51 & 52.12 & 52.43
 & 54.66 & 55.07 & \textbf{54.74}
 & \textbf{55.14} & 52.37 & 54.56
 & 54.67 & 53.13 & 54.35 \\
\midrule
\openaiicon~GPT-5.4
 & 42.85 & \textbf{58.44} & 45.26
 & 56.56 & 47.59 & 54.50
 & 56.38 & 51.11 & 55.24
 & \textbf{63.65} & 41.40 & \textbf{57.47}
 & 57.58 & 48.40 & 55.47 \\
\claudeicon~Claude Opus 4.6
 & 57.63 & \textbf{51.11} & 56.20
 & \textbf{70.16} & 30.74 & 55.84
 & 67.90 & 33.88 & 56.55
 & 66.13 & 40.62 & 58.75
 & 66.50 & 41.33 & \textbf{59.28} \\
\geminiicon~Gemini 3.1-Pro
 & 44.27 & \textbf{58.37} & 46.51
 & \textbf{60.14} & 48.61 & 57.41
 & 57.67 & 53.86 & 56.86
 & 59.48 & 49.69 & 57.22
 & 58.96 & 52.49 & \textbf{57.54} \\
\bottomrule
\end{tabular}
\caption{\textbf{Evaluation of prompting strategies on BEA-2019 dev}. Precision, recall, and $F_{0.5}$ scores for
the five manual prompts (ZS = zero-shot, FS = few-shot) across all seven models, evaluated at batch size $B{=}1$.
Per-model best metrics across prompts are bolded.}
\label{tab:taxonomy-prompts}
\end{table*}

\paragraph{Setup.}
We evaluate the five manually engineered prompts from Section~\ref{sec:prompting-strategies} on the BEA-2019 dev split
at batch size $B{=}1$. Precision, recall, and $F_{0.5}$ scores across our prompts are reported in
Table~\ref{tab:taxonomy-prompts}.

\paragraph{Results.}
The vanilla zero-shot prompt (\ref{prompt:vanilla-zero-shot}) shows a classic overcorrection pattern across all models,
with high recall but consistently poor precision (often below $45.0$). Imposing a minimal-edits constraint
(\ref{prompt:minimal-edits-zero-shot}) yields the largest gain, boosting $F_{0.5}$ by $+1.1$ to $+10.9$
points for all models except Claude Opus 4.6, where it is nearly neutral, but at the cost of a
$6$ to $22$ point drop in recall.

To recover this lost recall, we extend the minimal-edits strategy with in-context examples and the ERRANT taxonomy.
Few-shot prompting (\ref{prompt:minimal-edits-few-shot}) restores part of the lost recall for six of the
seven models and lifts $F_{0.5}$ by an additional $+1.3$ to $+3.0$ points across most models. The taxonomy's contribution, added on top of the few-shot
examples, is model-dependent rather than uniformly helpful. It improves two of the three high-capacity models,
Claude Opus 4.6 and Gemini 3.1-Pro, and degrades the remaining four commercial models, including the high-capacity GPT-5.4. The
taxonomy thus acts as a regularizer of the permissible edit space only for the strongest models, which appear to
read it as a strict constraint on admissible edits, and as noise for the others.
The open-weight Qwen3-8B follows the same qualitative pattern: the minimal-edits constraint trades recall
for precision, and few-shot examples yield its best $F_{0.5}$ ($48.98$), while the taxonomy provides no benefit.
Its best score trails all six commercial models, consistent with its much smaller scale.

While the optimal prompt structure varies across models, the absolute highest $F_{0.5}$ scores come from pairing our
strongest models with the minimal-edits few-shot + taxonomy prompt (\ref{prompt:taxonomy-few-shot}), peaking at
$59.28$ for Claude Opus 4.6 and $57.54$ for Gemini 3.1-Pro. Given its top-end performance, we adopt this prompt for
the batching experiments in Section~\ref{sec:exp-batching} and as the foundation for LLM-assisted Prompt Optimization
in Section~\ref{sec:exp-apo}.

\subsection{Batching Effect}
\label{sec:exp-batching}

\begin{figure*}[t!]
\centering
\includegraphics[width=\textwidth]{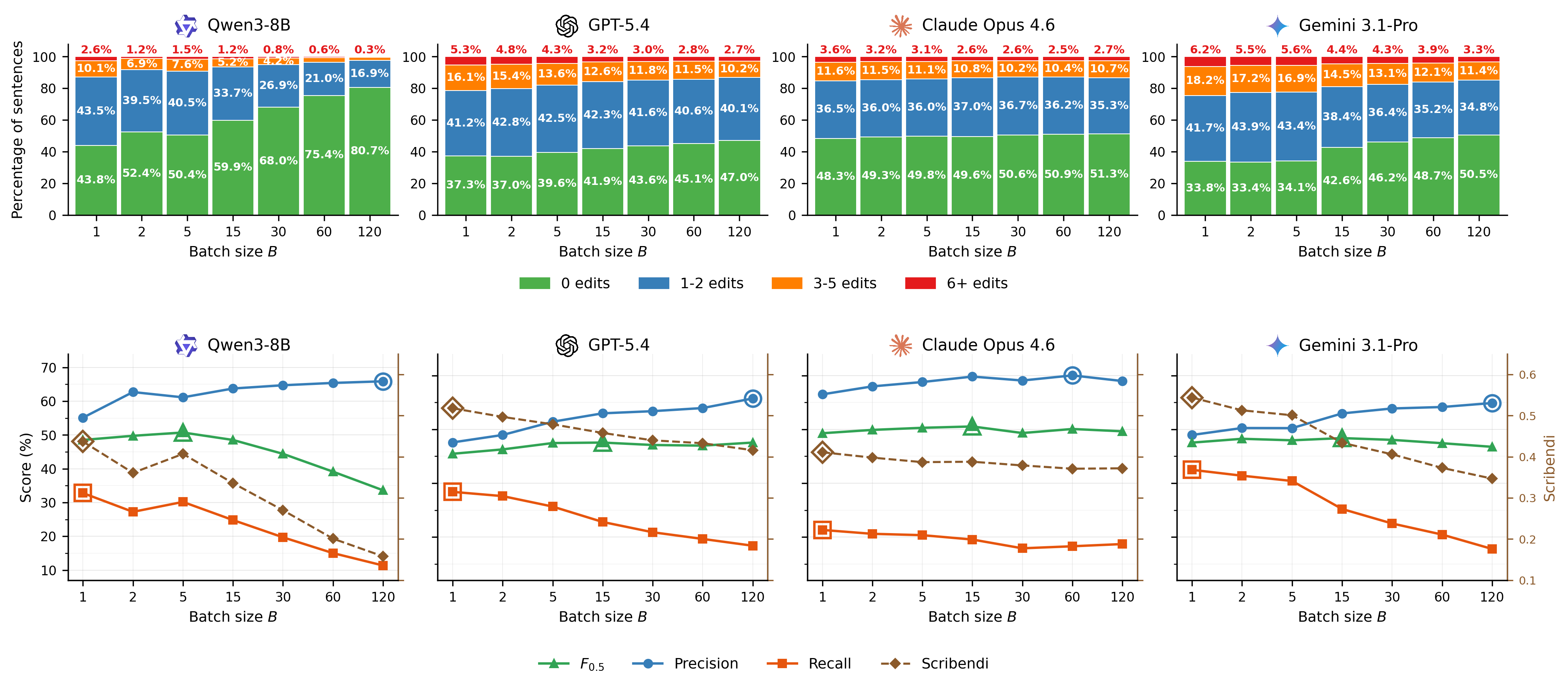}

\caption{\textbf{Edit-count distributions (top row) and GEC
metrics (bottom row) across batch sizes on
BEA-2019 dev under our best performing manual prompt: Minimal-edits few-shot + taxonomy (\ref{prompt:taxonomy-few-shot})}
Top row: per-sentence word-level edit counts as a function of the batch size $B$. As the batch size $B$ grows, the
general tendency is to shift mass into the 0-edit bucket (green), with the largest relative
reductions in the 6+ tail (red) and the 3--5 bucket (orange).
Bottom row: Precision (blue circles), Recall (orange squares), and $F_{0.5}$ (green triangles) on the
left axis, and the Scribendi score on the right axis (brown dashed diamonds), as a function of the batch size $B$.
Open rings mark the per-model Precision, Recall, $F_{0.5}$, and Scribendi maxima. See more results in Appendix~\ref{app:batching-results}.}

\label{fig:edit-distribution-best-prompt}
\label{fig:batching-trends-best-prompt}
\end{figure*}

\paragraph{Setup.} We vary batch sizes $B \in \{1, 2, 5, 15, 30, 60, 120\}$ on the BEA-2019 dev using the
minimal-edits few-shot + taxonomy prompt (\ref{prompt:taxonomy-few-shot}). We report per-sentence edit counts and
standard GEC metrics alongside the Scribendi fluency score
\citep{islam-magnani-2021-scribendi}.\footnote{Computed with Gemma 2 9B as the scoring LM following the MultiGEC-2025
protocol \citep{masciolini2025multigec}.}

\paragraph{Results.}
Scaling the batch size $B$ induces a systematic behavioral shift in Qwen3-8B and the three high-capacity models,
suggesting that
batching acts as a targeted regularizer at the per-sentence level
(Figure~\ref{fig:edit-distribution-best-prompt}, top row). As $B$ grows, probability mass moves steadily into the
0-edit bucket while the heavy-edit tail ($\geq 6$ edits) shrinks, suppressing the extreme rewrites typically
associated with stylistic or hallucinated output. The shift is most pronounced for the open-weight Qwen3-8B, whose
share of untouched sentences nearly doubles from $43.8\%$ at $B{=}1$ to $80.7\%$ at $B{=}120$, and more moderate
for the high-capacity models (e.g., $33.8\%$ to $50.5\%$ for Gemini 3.1-Pro).

This conservative shift is mirrored in the standard GEC metrics
(Figure~\ref{fig:batching-trends-best-prompt}, bottom row). Precision rises with the batch size for all four
models, peaking at the largest batches, while recall moves in the opposite direction. Claude Opus 4.6, for
instance, reaches the highest precision of the sweep, $70.02$ at $B{=}60$. Because $F_{0.5}$ weights precision heavily, the two
trends largely offset each other for the three high-capacity models, producing a flat profile with a shallow
optimum at the moderate batch size $B{=}15$; for Qwen3-8B the optimum comes earlier, at $B{=}5$.

Batching also carries a qualitative cost, captured by the reference-less Scribendi fluency score. For all four
models, the score peaks at single-sentence prompting ($B{=}1$) and declines nearly monotonically as the
batch size grows: in a strictly minimal-edit setting, large batches heavily discourage overcorrection, so the
models leave progressively more disfluencies intact.

The three medium-capacity models largely follow the same trends, with their $F_{0.5}$ optima shifted toward larger
batches ($B{=}30$ to $B{=}60$); complete per-model curves and numerical results for all seven models are provided
in Appendix~\ref{app:batching-results} (Figure~\ref{fig:edit-distribution-best-prompt-6},
Tables~\ref{tab:batching-best-prompt} and~\ref{tab:edit-distribution-best-prompt}). The batching effect thus
generalizes across capacity tiers.

\textbf{Hypothesis: Batching limits the attention budget.} While internal LLM dynamics remain opaque, we conjecture
that this conservative shift is driven by an \textit{attention dilution} effect. Because the self-attention softmax
enforces a normalization constraint, packing independent sentences into a single prompt limits the
attention budget per token. At batch size $B = 1$, models have enough focus to perform stylistic rewrites;
as batch size $B$ grows, a flattened attention distribution forces the model to prioritize, pivoting away from
adjustments and committing to core grammatical errors where the error signal remains salient.

\textbf{Practical implications of batching.} The quality-optimal batch size for the three high-capacity models is
$B = 15$ (Figure~\ref{fig:batching-trends-best-prompt}). Such a batch is a realistic unit of work: a single A4 page holds roughly 20 to 40
sentences, and proofreading services, learner-platform analytics, and corpus-cleaning pipelines routinely submit
many independent sentences per request.

For genuinely single-sentence requests, a promising direction is to pad the
target sentence with auxiliary erroneous sentences whose corrections are discarded, analogous to few-shot
demonstrations but without reference outputs, which would make the gain observed at $B = 15$ available even when
only one sentence needs corrections.

\subsection{LLM-assisted Prompt Optimization}
\label{sec:exp-apo}

\begin{table*}[t]
\centering
\scriptsize
\setlength{\tabcolsep}{2pt}
\begin{tabular}{l c @{\hskip 6pt} rrr @{\hskip 10pt} rrr @{\hskip 10pt} rrr @{\hskip 10pt} rrr}
\toprule
 & & \multicolumn{3}{c}{\qwenicon~Qwen3-8B}
 & \multicolumn{3}{c}{\openaiicon~GPT-4.1-mini}
 & \multicolumn{3}{c}{\kern-6pt\claudeicon~Claude Sonnet 4.6\kern-6pt}
 & \multicolumn{3}{c}{\geminiicon~Gemini 3-Flash} \\
\cmidrule(lr){3-5}\cmidrule(lr){6-8}\cmidrule(lr){9-11}\cmidrule(lr){12-14}
\textbf{Prompt} & $\mathbf{B}$ & \textbf{Prec.} & \textbf{Rec.} & $\mathbf{F_{0.5}}$
 & \textbf{Prec.} & \textbf{Rec.} & $\mathbf{F_{0.5}}$
 & \textbf{Prec.} & \textbf{Rec.} & $\mathbf{F_{0.5}}$
 & \textbf{Prec.} & \textbf{Rec.} & $\mathbf{F_{0.5}}$ \\
\midrule
Minimal-edits few-shot + taxonomy (\ref{prompt:taxonomy-few-shot}) & 5
 & 61.15 & 30.18 & 50.74
 & -- & -- & --
 & -- & -- & --
 & -- & -- & -- \\
Minimal-edits few-shot + taxonomy (\ref{prompt:taxonomy-few-shot}) & 15
 & -- & -- & --
 & 61.02 & 41.58 & 55.80
 & 65.02 & 39.46 & 57.56
 & 55.79 & 53.12 & 55.23 \\
\midrule
\quad + optimized for Qwen3-8B (\ref{prompt:apo:qwen}) & 5
 & 60.65\rlap{$^{\dagger}$} & \textbf{33.68}\rlap{$^{\dagger}$} & \textbf{52.28}\rlap{$^{\dagger}$}
 & 59.79 & 41.41 & 54.92
 & 67.69 & 37.96 & 58.52
 & 58.04 & 51.72 & 56.66 \\
\quad + optimized for GPT-4.1-mini (\ref{prompt:apo:gpt}) & 15
 & 63.61 & 18.02 & 42.24
 & 62.98\rlap{$^{\dagger}$} & 42.39\rlap{$^{\dagger}$} & \textbf{57.40}\rlap{$^{\dagger}$}
 & 67.14 & 37.64 & 58.04
 & 61.22 & 48.50 & 58.17 \\
\quad + optimized for Claude Sonnet~4.6 (\ref{prompt:apo:claude}) & 15
 & \textbf{64.16} & 21.05 & 45.52
 & 61.84 & \textbf{43.31} & 56.97
 & 66.88\rlap{$^{\dagger}$} & \textbf{41.21}\rlap{$^{\dagger}$} & \textbf{59.47}\rlap{$^{\dagger}$}
 & 59.53 & \textbf{53.50} & 58.22 \\
\quad + optimized for Gemini~3-Flash (\ref{prompt:apo:gemini}) & 15
 & 61.45 & 6.83 & 23.64
 & \textbf{65.73} & 31.73 & 54.13
 & \textbf{68.52} & 36.64 & 58.36
 & \textbf{65.45}\rlap{$^{\dagger}$} & 42.41\rlap{$^{\dagger}$} & \textbf{59.04}\rlap{$^{\dagger}$} \\
\bottomrule
\end{tabular}
\caption{\textbf{Evaluation of optimized prompts across different models on the BEA-2019 development set.}
 Each prompt is specifically optimized using one model and
subsequently evaluated on all four models without further modification, at the batch size given in
 the $B$ column.
 The highest Precision,
 Recall, and $F_{0.5}$ values within each column are highlighted in bold.
 Instances of native evaluation, indicating that the prompt is tested on its respective tuning model, are
 indicated with $^{\dagger}$ across all metrics.}
\label{tab:apo_transfer}
\end{table*}

\paragraph{Setup.}

Initializing with our comprehensive minimal-edits few-shot + taxonomy prompt
(\ref{prompt:taxonomy-few-shot}), we obtain four model family-specific prompts optimized for
the open-weight Qwen3-8B (\ref{prompt:apo:qwen}), OpenAI GPT (\ref{prompt:apo:gpt}), Anthropic Claude
(\ref{prompt:apo:claude}), and Google Gemini (\ref{prompt:apo:gemini}). We restricted the tuning phase to
the open-weight and medium-capacity models because of budget limitations. The prompt selection was done based
on $F_{0.5}$ performance on the BEA-2019 dev after assessing approximately one hundred candidates
per model. As a sanity check of robustness, we evaluate each optimized prompt not only on its tuning
model but on all four models: Table~\ref{tab:apo_transfer} details their performance at a batch size of
$B=15$; the Qwen3-8B-optimized prompt (\ref{prompt:apo:qwen}) is optimized and
evaluated at $B=5$, the batch size at which Qwen3-8B performs best (Figure~\ref{fig:batching-trends-best-prompt}).

\paragraph{Results.}
The optimization procedure enhances performance considerably: as anticipated, the gains are largest under
native evaluation, where the prompt was selected on the same model, yet they often transfer effectively to other
architectures. For instance, the prompt optimized for Claude Sonnet 4.6 (\ref{prompt:apo:claude}) increases
the $F_{0.5}$ score on the GPT-4.1-mini model from $55.80$ to $56.97$ and elevates the Gemini performance from $55.23$
to $58.22$. Similarly, the Gemini-optimized instruction (\ref{prompt:apo:gemini}) improves the Claude model
score from $57.56$ to $58.36$. The open-weight Qwen3-8B follows the same native-is-best pattern: its own
optimized prompt (\ref{prompt:apo:qwen}) raises its $F_{0.5}$ from $50.74$ to $52.28$, whereas the prompts tuned for the
proprietary models transfer poorly to it, with the Gemini-optimized prompt (\ref{prompt:apo:gemini}) collapsing its
recall and dropping $F_{0.5}$ to $23.64$. Since native evaluation remains optimal despite this positive transfer, in Section~\ref{sec:exp-test} we
pair each medium-capacity optimized prompt with its high-capacity sibling from the same model family, leaving a
deeper study of cross-family transferability to future work.

\subsection{Final Evaluations}
\label{sec:exp-test}

\begin{table*}[t]
\centering
\scriptsize
\setlength{\tabcolsep}{2pt}
\begin{tabular}{ll c rrr @{\hskip 6pt} rrr}
\toprule
 & & & \multicolumn{3}{c}{\textbf{CoNLL-2014 test}} & \multicolumn{3}{c}{\textbf{BEA-2019 test}} \\
\cmidrule(lr){4-6}\cmidrule(lr){7-9}
\textbf{Prompt} & \textbf{Model} & $\mathbf{B}$ & \textbf{Prec.} & \textbf{Rec.} & $\mathbf{F_{0.5}}$
 & \textbf{Prec.} & \textbf{Rec.} & $\mathbf{F_{0.5}}$ \\
\midrule
\multicolumn{9}{l}{\textit{Single-system SOTA}} \\
\midrule
Fine-tuned, EPO \citep{liang2025edit} & \mistralicon~Mistral (7B) & --
 & 76.71 & \textbf{52.56} & \textbf{70.26} & 78.16 & \textbf{68.07} & 75.91 \\
Fine-tuned, Training Schedule \citep{staruch2025adapting} & \gemmaicon~Gemma 2 (27B) & --
 & \textbf{77.38} & 47.88 & 68.89 & \textbf{82.28} & 67.03 & \textbf{78.70} \\
\midrule
\multicolumn{9}{l}{\textit{LLM-prompted baselines}} \\
\midrule
3-/1-shot CoT \citep{fang2023chatgpt} & \openaiicon~GPT-3.5-Turbo-0301 & --
 & 51.3 & \textbf{62.4} & 53.2 & 34.6 & 69.7 & 38.4 \\
16-shot \citep{loem2023exploring} & \openaiicon~GPT-3.5 text-davinci-003 & --
 & -- & -- & 57.06 & -- & -- & 57.41 \\
2-shot \citep{coyne2023analyzing} & \openaiicon~GPT-4-0314 & --
 & -- & -- & -- & -- & -- & 52.79 \\
1-shot \citep{davis2024prompting} & \openaiicon~GPT-3.5-Turbo-0613 & --
 & -- & -- & 57.2 & -- & -- & -- \\
0-shot \citep{chernodub2025apio} & \openaiicon~GPT-4o & --
 & -- & -- & -- & -- & -- & 59.40 \\
0-shot minimal-edit \citep{turker-eryigit-2026-instruction} & \claudeicon~Claude Sonnet 4.5
 & -- & \textbf{69.14} & 59.82 & \textbf{67.05} & 63.48
 & \textbf{71.33} & 64.91 \\
4-shot \citep{goto-etal-2026-edit} & \qwenicon~Qwen3-8B (majority voting on 8 runs) & --
 & -- & -- & -- & \textbf{73.8} & 53.7 & \textbf{68.7} \\
\midrule
\multicolumn{9}{l}{\textit{Our results}} \\
\midrule
Vanilla zero-shot (\ref{prompt:vanilla-zero-shot}) & \multirow{3}{*}{\qwenicon~Qwen3-8B} & 1
 & 54.11 & \textbf{49.49} & 53.12 & 54.11 & \textbf{65.85} & 56.11 \\
Minimal-edits few-shot + taxonomy (\ref{prompt:taxonomy-few-shot}) & & 5
 & 71.54 & 28.52 & 54.95 & \textbf{73.85} & 50.64 & 67.65 \\
Minimal-edits few-shot + taxonomy + optimized (\ref{prompt:apo:qwen}) & & 5
 & \textbf{71.74} & 35.83 & \textbf{59.76} & 72.14 & 58.21 & \textbf{68.84} \\
\midrule
Vanilla zero-shot (\ref{prompt:vanilla-zero-shot}) & \multirow{3}{*}{\openaiicon~GPT-4.1-mini} & 1
 & 50.74 & 57.91 & 52.03 & 54.17 & 73.21 & 57.14 \\
Minimal-edits few-shot + taxonomy (\ref{prompt:taxonomy-few-shot}) & & 60
 & 74.24 & 45.30 & \textbf{65.83} & 76.14 & 64.87 & 73.59 \\
Minimal-edits few-shot + taxonomy + optimized (\ref{prompt:apo:gpt}) & & 60
 & \textbf{74.30} & 43.66 & 65.15 & 76.70 & 64.24 & 73.83 \\
\lightmodelrule
Vanilla zero-shot (\ref{prompt:vanilla-zero-shot}) & \multirow{3}{*}{\claudeicon~Claude Sonnet 4.6} & 1
 & 53.47 & 57.87 & 54.30 & 59.61 & 72.05 & 61.74 \\
Minimal-edits few-shot + taxonomy (\ref{prompt:taxonomy-few-shot}) & & 60
 & 73.12 & 44.43 & 64.75 & 76.13 & 64.62 & 73.51 \\
Minimal-edits few-shot + taxonomy + optimized (\ref{prompt:apo:claude}) & & 60
 & 72.51 & 44.37 & 64.35 & \textbf{77.32} & 65.36 & \textbf{74.59} \\
\lightmodelrule
Vanilla zero-shot (\ref{prompt:vanilla-zero-shot}) & \multirow{3}{*}{\geminiicon~Gemini 3 Flash} & 1
 & 46.52 & \textbf{59.92} & 48.70 & 51.99 & 73.22 & 55.19 \\
Minimal-edits few-shot + taxonomy (\ref{prompt:taxonomy-few-shot}) & & 30
 & 62.03 & 56.83 & 60.91 & 67.65 & \textbf{73.31} & 68.71 \\
Minimal-edits few-shot + taxonomy + optimized (\ref{prompt:apo:gemini}) & & 30
 & 70.48 & 49.31 & 64.91 & 74.94 & 68.98 & 73.67 \\
\midrule
Vanilla zero-shot (\ref{prompt:vanilla-zero-shot}) & \multirow{3}{*}{\openaiicon~GPT-5.4} & 1
 & 45.82 & 60.56 & 48.17 & 51.37 & 73.68 & 54.68 \\
Minimal-edits few-shot + taxonomy (\ref{prompt:taxonomy-few-shot}) & & 15
 & 69.20 & 49.17 & 63.99 & 74.85 & 67.73 & 73.31 \\
Minimal-edits few-shot + taxonomy + optimized (\ref{prompt:apo:gpt}) & & 15
 & 72.63 & 47.10 & 65.53 & 77.62 & 63.65 & 74.36 \\
\lightmodelrule
Vanilla zero-shot (\ref{prompt:vanilla-zero-shot}) & \multirow{3}{*}{\claudeicon~Claude Opus 4.6} & 1
 & 62.57 & 54.50 & 60.77 & 68.77 & 71.33 & 69.27 \\
Minimal-edits few-shot + taxonomy (\ref{prompt:taxonomy-few-shot}) & & 15
 & 74.76 & 45.05 & 66.04 & 78.58 & 63.94 & 75.14 \\
Minimal-edits few-shot + taxonomy + optimized (\ref{prompt:apo:claude}) & & 15
 & 74.90 & 44.27 & 65.80 & 80.72 & 64.05 & 76.73 \\
\lightmodelrule
Vanilla zero-shot (\ref{prompt:vanilla-zero-shot}) & \multirow{3}{*}{\geminiicon~Gemini 3.1-Pro} & 1
 & 45.94 & \textbf{60.81} & 48.30 & 53.32 & \textbf{73.81} & 56.45 \\
Minimal-edits few-shot + taxonomy (\ref{prompt:taxonomy-few-shot}) & & 15
 & 66.72 & 50.17 & 62.59 & 74.13 & 68.24 & 72.87 \\
Minimal-edits few-shot + taxonomy + optimized (\ref{prompt:apo:gemini}) & & 15
 & \textbf{76.20} & 45.36 & \textbf{67.08} & \textbf{82.29} & 65.66 & \textbf{78.32} \\
\bottomrule
\end{tabular}
\caption{\textbf{Final evaluations on CoNLL-2014 test and BEA-2019 test.} Precision, recall, and $F_{0.5}$
on all considered models under three prompt configurations: vanilla zero-shot at
batch size $B{=}1$, minimal-edits few-shot + taxonomy at the per-model dev-optimal batch size (Figure \ref{fig:batching-trends-best-prompt}), and its
optimized variant at the same per-model batch size. Within each group, the best precision, recall, and
$F_{0.5}$ per dataset are bolded.}
\label{tab:final-results}
\end{table*}

\paragraph{Experimental Setup.}
We evaluate three prompt configurations on the held out CoNLL 2014 and BEA 2019 test sets across all
considered models. Configurations include a vanilla zero-shot prompt (\ref{prompt:vanilla-zero-shot}) at
$B{=}1$, a minimal edits few-shot + taxonomy prompt (\ref{prompt:taxonomy-few-shot}), and its
model-specific optimized variants (\ref{prompt:apo:qwen}--\ref{prompt:apo:gemini}), both
evaluated at each model's dev-optimal batch size (selected by $F_{0.5}$ on the BEA-2019 dev
split). Table~\ref{tab:final-results} contextualizes
performance of our models against leading prompted and fine-tuned baselines.

\paragraph{Results.}
Empirical results show that batched inference coupled with strict taxonomy constraints jointly help to mitigate
the overcorrection. Vanilla configurations universally exhibit high recall but poor
precision. Introducing taxonomy rules and batched generation restricts extraneous edits, substantially shifting
performance toward the precision weighted $F_{0.5}$ metric. Notably, Gemini 3.1-Pro improves its BEA-2019 precision by $20.81$ points upon adding these joint
constraints (\ref{prompt:taxonomy-few-shot}), yielding a $16.4$ point
$F_{0.5}$ gain. The medium capacity tier exhibits similar absolute $F_{0.5}$ improvements of over $16$ points for
GPT-4.1-mini and roughly $13$ points for Gemini 3 Flash, confirming that structural framing consistently aids
precision recovery.

LLM assisted prompt optimization provides additional model-specific advantages. The prompt optimized for
Gemini 3.1-Pro (\ref{prompt:apo:gemini}) establishes
a new state of the art for prompt-based GEC, outperforming the strongest prompted baseline \citep{goto-etal-2026-edit}
by nearly $10$ $F_{0.5}$ points on BEA-2019. Crucially, this framework achieves parity with state-of-the-art
fine-tuned paradigms. Our optimal high-capacity configuration matches the precision of the leading fine-tuned
model \citep{staruch2025adapting} on BEA-2019 ($82.29$ vs. $82.28$) and trails by a marginal $0.38$ $F_{0.5}$ points,
effectively tying the current fine-tuned SOTA. Furthermore, on BEA-2019 our optimized medium-capacity models surpass all existing
prompted baselines, demonstrating the efficiency of structured constraints. The zero-shot minimal-edit prompt of \citet{turker-eryigit-2026-instruction} on Claude Sonnet 4.5
closely resembles our baseline (\ref{prompt:minimal-edits-zero-shot}); on Claude Sonnet 4.6, adding few-shot
examples, taxonomy, batching, and prompt optimization raises $F_{0.5}$ to $74.59$ on BEA-2019, $9.68$ points above
their result.

The open-weight Qwen3-8B follows the
same pattern at a smaller scale: its optimized prompt (\ref{prompt:apo:qwen}) at $B{=}5$ reaches
$F_{0.5}=68.84$ on BEA-2019 test, a $12.73$-point gain over its vanilla zero-shot configuration that edges out the
strongest Qwen-based prompted baseline \citep{goto-etal-2026-edit} ($68.7$) without any self-ensembling: a single run
matches the quality that the baseline obtains by majority voting over 8 runs.

At the same time, we observe a cross domain
generalization gap on the CoNLL-2014 set for two of the three medium-capacity models: prompt optimization
lowers $F_{0.5}$ by $0.68$ points for GPT-4.1-mini and by $0.40$ points for Claude Sonnet 4.6.

\section{Conclusion}
We present a prompt-based framework for minimal-edit GEC that bridges the gap between off-the-shelf LLM prompting
and state-of-the-art fine-tuned single systems. Its core technique, batched inference layered on minimal-edit
instructions, acts as a targeted regularizer against overcorrection, systematically reducing the edit rate
across diverse LLM families, while simultaneously lowering inference cost and latency. The taxonomy
component, by contrast, proves model-dependent: it constrains the edit space productively only for the strongest
models, a finding that motivates our final component, per-model prompt optimization.

We evaluate our approach on six commercial and one open-weight LLM across two minimal-edit benchmarks. The
resulting prompts set a new state of the art for prompt-based minimal-edit GEC: with Gemini 3.1-Pro we reach
$F_{0.5}=78.32$ on the BEA-2019 test set, within $0.38$ points of the best fine-tuned single model --- to our
knowledge, the first purely prompt-based system to reach parity with the fine-tuned state of the art in
GEC --- alongside a competitive $F_{0.5}=67.08$ on CoNLL-2014.

These results show that careful prompt design and context management let general-purpose
LLMs perform minimal-edit GEC at fine-tuned quality without the computational overhead of fine-tuning, albeit at
the cost of a conservative editing style that may limit fluency-oriented rewriting. This makes high-quality GEC
practical for settings where fine-tuned models cannot be deployed for resource or regulatory reasons.
Our framework\footref{fn:repo} can also generate high-quality synthetic training data for fine-tuning GEC models.

\section*{Limitations}
Our study has several limitations that frame the scope of its conclusions and motivate concrete avenues for future work.

\begin{enumerate}
\item While the BEA-2019 \citep{bryant2019bea} test set is held-out on CodaBench
platform\footnote{\url{https://www.codabench.org/competitions/10960/}}, its public train and dev splits, along with the
fully public CoNLL-2014 \citep{ng2014conll} dataset, may have been encountered by commercial LLMs during pretraining.
Exposure to these specific domains and annotation styles could still inflate performance. Consequently, our
results may establish strong prompt-based baselines rather than strict measures of generalization to completely unseen text.

\item We rely mostly on commercial APIs, making weights, training data, and inference stacks opaque, and leaving us vulnerable
to silent provider updates. Furthermore, due to costs, each configuration is evaluated only once without reporting
variance, meaning exact reproducibility is not strictly guaranteed and
broader systematic comparisons with open-weight models, beyond our single open-weight baseline, are left to
future work.

\item Our LLM-assisted Prompt Optimization (Section~\ref{sec:apo}) tunes prompts only on medium-capacity tier
models (GPT-4.1-mini, Claude Sonnet 4.6, Gemini 3 Flash) and the open-weight Qwen3-8B due to budget
constraints, and does not tune on the high-capacity
siblings (GPT-5.4, Claude Opus 4.6, Gemini 3.1-Pro). This leaves open the question of whether optimization on
high-capacity models would yield similar or superior prompts, and how these might transfer across model families.

\item Evaluations are restricted to English learner essays. It remains unclear how our
taxonomy instructions and batching transfer to morphologically richer languages, lower-resource settings, professional
editing, or fluency-oriented benchmarks like JFLEG \citep{napoles2017jfleg}.

\item We optimize and evaluate primarily on precision-weighted $F_{0.5}$ computed by ERRANT
and the $M^{2}$ scorer. While this aligns with minimal-edit goals, it does not capture fluency or stylistic
improvements beyond the reference-free Scribendi score used in the batching analysis
(Section~\ref{sec:exp-batching}). We omit human evaluation and GLEU scores, so our findings represent only a
precision-weighted slice of GEC performance.

\item The claim that batching regularizes via self-attention dilution
(Section~\ref{sec:exp-batching}) is an empirical conjecture based on systematic trends across batch sizes.
Confirming this rigorously requires controlled probes on open-weight models, which we do not perform.

\item Some of our LLM-optimized prompts are significantly longer
(Figure~\ref{fig:prompt-tokens}), increasing inference token costs and latency. Additionally, real-world
deployment involves topical streams rather than the shuffled independent sentences used in our analysis,
which could alter attention dynamics and introduce unquantified generation failures in the structured response.

\end{enumerate}

\section*{Ethical Considerations}
We disclose that ChatGPT, Claude, Gemini, and  Grammarly were used for drafting, editing, and proofreading.
All AI-generated text was reviewed by the authors, who take full responsibility for the final content.

\bibliography{custom}

\clearpage
\appendix

\section{Prompts}
\label{sec:appendix-prompts}

Below we list all prompts used in our experiments. The placeholder \texttt{<input\_text>} is replaced with the input
sentence (or the batched block of sentences) at inference time. All prompts target
English minimal-edit GEC.

\subsection{Vanilla zero-shot prompt}
\label{prompt:vanilla-zero-shot}

\begin{promptbox}
\small
\texttt{Reply with a corrected version of the input sentence with all grammatical and spelling errors fixed.}\\
\texttt{If there are no errors, reply with a copy of the original sentence.}
\end{promptbox}

\subsection{Minimal-edits zero-shot prompt}
\label{prompt:minimal-edits-zero-shot}

\begin{promptbox}
\small
\texttt{You are a grammatical error correction system. Make MINIMAL, PRECISE edits to fix errors.}\\
\texttt{DO NOT rewrite or paraphrase. Only fix clear grammatical and spelling errors.}\\[1mm]
\texttt{RULES:}\\
\texttt{- Make the SMALLEST possible edit to fix each error}\\
\texttt{- Change only what is grammatically or orthographically wrong}\\
\texttt{- Preserve the original meaning and style}\\
\texttt{- Do NOT improve fluency beyond fixing errors}\\
\texttt{- Do NOT make stylistic changes}\\
\texttt{- If no errors exist, return the original sentence unchanged}
\end{promptbox}

\subsection{Minimal-edits few-shot prompt}
\label{prompt:minimal-edits-few-shot}

\begin{promptbox}
\small
\texttt{You are a grammatical error correction system. Make MINIMAL, PRECISE edits to fix errors.}\\
\texttt{DO NOT rewrite or paraphrase. Only fix clear grammatical and spelling errors.}\\[1mm]
\texttt{RULES:}\\
\texttt{- Make the SMALLEST possible edit to fix each error}\\
\texttt{- Change only what is grammatically or orthographically wrong}\\
\texttt{- Preserve the original meaning and style}\\
\texttt{- Do NOT improve fluency beyond fixing errors}\\
\texttt{- Do NOT make stylistic changes}\\
\texttt{- If no errors exist, return the original sentence unchanged}\\[1mm]
\texttt{Few-shot examples from BEA train:}\\
\texttt{Follow the same input-to-correction style. Do not copy these examples; use them only as guidance.}\\[1mm]
\texttt{Example 1:}\\
\texttt{Input: Alison put on her coat, close the door and went to his parents house by car.}\\
\texttt{Correction: Alison put on her coat, closed the door and went to her parents ' house by car.}\\[1mm]
\texttt{Example 2:}\\
\texttt{Input: we can see the number of the projected sales of jeans of Jack\&Jones Co. will be at 450 thousands of pairs in next Jan, it will be higher than the number of the projected sales of jeans of Mango Co. about 300 thousands of pairs.Then the number of Jack\&Jones Co. will drop to 250 thousands of pairs in Feb next year.}\\
\texttt{Correction: we can see the number of projected sales of jeans of Jack\&Jones Co. will be at 450 thousand pairs next Jan, it will be higher than the number of the projected sales of jeans of Mango Co.; about 300 thousand pairs. Then the number of Jack\&Jones Co. will drop to 250 thousand pairs in Feb next year.}\\[1mm]
\texttt{Example 3:}\\
\texttt{Input: The girl is 18 years old and men (as it later turns out) are vampires and they have over 100 years.}\\
\texttt{Correction: The girl is 18 years old and the men (as it later turns out) are vampires and they are over 100 years old.}\\[1mm]
\texttt{Example 4:}\\
\texttt{Input: Like what was once, I have no doubt that this world of wonders will remain important in the coming years.}\\
\texttt{Correction: Like what was once, I have no doubt that this world of wonders will remain important in the coming years.}\\[1mm]
\texttt{Example 5:}\\
\texttt{Input: The need of export and import items has increased in the last 10 years consequently the life of the people has raised in faster respond.}\\
\texttt{Correction: The need for exporting and importing items has increased in the last 10 years. Consequently, the life of the people has been raised in rapid response.}\\[1mm]
\texttt{Example 6:}\\
\texttt{Input: In conclusion, it's very important to learn the concept than saying something stupid....}\\
\texttt{Correction: In conclusion, it's more important to learn the concept than say saying something stupid....}\\[1mm]
\texttt{Example 7:}\\
\texttt{Input: Abortion is an immoral act that people make as everyone has a chance to survive and nobody has the right to decide for anyone life.}\\
\texttt{Correction: Abortion is an immoral act that people perform as everyone has the chance to survive and nobody has the right to decide about anyone's life.}\\[1mm]
\texttt{Example 8:}\\
\texttt{Input: Hi John:}\\
\texttt{Correction: Hi John:}
\end{promptbox}

\subsection{Minimal-edits zero-shot + taxonomy prompt}
\label{prompt:taxonomy-zero-shot}

\begin{promptbox}
\small
\texttt{You are a grammatical error correction system. Make MINIMAL, PRECISE edits to fix errors.}\\
\texttt{DO NOT rewrite or paraphrase. Only fix clear grammatical and spelling errors.}\\[1mm]
\texttt{Focus on these 25 error types:}\\[1mm]
\texttt{WORD-LEVEL ERRORS:}\\
\texttt{1. ADJ: Wrong adjective choice (big$\to$wide)}\\
\texttt{2. ADJ:FORM: Adjective form errors - comparatives/superlatives (goodest$\to$best,}\\
\texttt{more easy$\to$easier)}\\
\texttt{3. ADV: Wrong adverb choice (speedily$\to$quickly)}\\
\texttt{4. CONJ: Wrong conjunction (and$\to$but)}\\
\texttt{5. CONTR: Contraction errors (n't$\to$not)}\\
\texttt{6. DET: Wrong/missing/extra determiner (the$\to$a, $\emptyset\to$the, the$\to\emptyset$)}\\
\texttt{7. NOUN: Wrong noun choice (person$\to$people)}\\
\texttt{8. NOUN:INFL: Count-mass noun errors (informations$\to$information)}\\
\texttt{9. NOUN:NUM: Noun number agreement (cat$\to$cats)}\\
\texttt{10. NOUN:POSS: Noun possessive errors (friends$\to$friend's)}\\
\texttt{11. PART: Wrong particle (look in$\to$look at)}\\
\texttt{12. PREP: Wrong/missing/extra preposition (of$\to$at, $\emptyset\to$at, at$\to\emptyset$)}\\
\texttt{13. PRON: Wrong pronoun (ours$\to$ourselves)}\\
\texttt{14. VERB: Wrong verb choice (ambulate$\to$walk)}\\
\texttt{15. VERB:FORM: Verb form errors - infinitive/gerund/participle (to eat$\to$eating,}\\
\texttt{dancing$\to$danced)}\\
\texttt{16. VERB:INFL: Verb inflection errors (getted$\to$got, fliped$\to$flipped)}\\
\texttt{17. VERB:SVA: Subject-verb agreement ((He) have$\to$(He) has)}\\
\texttt{18. VERB:TENSE: Verb tense errors including modals and passive (eats$\to$ate, eats$\to$can eat,}\\
\texttt{eats$\to$was eaten)}\\[1mm]
\texttt{MECHANICAL ERRORS:}\\
\texttt{19. ORTH: Orthography - capitalization/whitespace (Bestfriend$\to$best friend, THIS$\to$this)}\\
\texttt{20. PUNCT: Punctuation errors (!$\to$., missing commas, extra periods)}\\
\texttt{21. SPELL: Spelling errors (genectic$\to$genetic, color$\to$colour)}\\
\texttt{22. WO: Word order errors (only can$\to$can only)}\\[1mm]
\texttt{OTHER:}\\
\texttt{23. MORPH: Morphology - same lemma, different part of speech (quick[adj]$\to$quickly[adv])}\\
\texttt{24. OTHER: Complex errors requiring minimal paraphrasing}\\
\texttt{25. UNK: Leave unchanged if error is unclear}\\[1mm]
\texttt{RULES:}\\
\texttt{- Make the SMALLEST possible edit to fix each error}\\
\texttt{- Change only what is grammatically or orthographically wrong}\\
\texttt{- Preserve the original meaning and style}\\
\texttt{- Do NOT improve fluency beyond fixing errors}\\
\texttt{- Do NOT make stylistic changes}\\
\texttt{- If no errors exist, return the original sentence unchanged}
\end{promptbox}

\subsection{Minimal-edits few-shot + taxonomy prompt}
\label{prompt:taxonomy-few-shot}

\begin{promptbox}
\small
\texttt{You are a grammatical error correction system. Make MINIMAL, PRECISE edits to fix errors.}\\
\texttt{DO NOT rewrite or paraphrase. Only fix clear grammatical and spelling errors.}\\[1mm]
\texttt{Focus on these 25 error types:}\\[1mm]
\texttt{WORD-LEVEL ERRORS:}\\
\texttt{1. ADJ: Wrong adjective choice (big$\to$wide)}\\
\texttt{2. ADJ:FORM: Adjective form errors - comparatives/superlatives (goodest$\to$best,}\\
\texttt{more easy$\to$easier)}\\
\texttt{3. ADV: Wrong adverb choice (speedily$\to$quickly)}\\
\texttt{4. CONJ: Wrong conjunction (and$\to$but)}\\
\texttt{5. CONTR: Contraction errors (n't$\to$not)}\\
\texttt{6. DET: Wrong/missing/extra determiner (the$\to$a, $\emptyset\to$the, the$\to\emptyset$)}\\
\texttt{7. NOUN: Wrong noun choice (person$\to$people)}\\
\texttt{8. NOUN:INFL: Count-mass noun errors (informations$\to$information)}\\
\texttt{9. NOUN:NUM: Noun number agreement (cat$\to$cats)}\\
\texttt{10. NOUN:POSS: Noun possessive errors (friends$\to$friend's)}\\
\texttt{11. PART: Wrong particle (look in$\to$look at)}\\
\texttt{12. PREP: Wrong/missing/extra preposition (of$\to$at, $\emptyset\to$at, at$\to\emptyset$)}\\
\texttt{13. PRON: Wrong pronoun (ours$\to$ourselves)}\\
\texttt{14. VERB: Wrong verb choice (ambulate$\to$walk)}\\
\texttt{15. VERB:FORM: Verb form errors - infinitive/gerund/participle (to eat$\to$eating,}\\
\texttt{dancing$\to$danced)}\\
\texttt{16. VERB:INFL: Verb inflection errors (getted$\to$got, fliped$\to$flipped)}\\
\texttt{17. VERB:SVA: Subject-verb agreement ((He) have$\to$(He) has)}\\
\texttt{18. VERB:TENSE: Verb tense errors including modals and passive (eats$\to$ate, eats$\to$can eat,}\\
\texttt{eats$\to$was eaten)}\\[1mm]
\texttt{MECHANICAL ERRORS:}\\
\texttt{19. ORTH: Orthography - capitalization/whitespace (Bestfriend$\to$best friend, THIS$\to$this)}\\
\texttt{20. PUNCT: Punctuation errors (!$\to$., missing commas, extra periods)}\\
\texttt{21. SPELL: Spelling errors (genectic$\to$genetic, color$\to$colour)}\\
\texttt{22. WO: Word order errors (only can$\to$can only)}\\[1mm]
\texttt{OTHER:}\\
\texttt{23. MORPH: Morphology - same lemma, different part of speech (quick[adj]$\to$quickly[adv])}\\
\texttt{24. OTHER: Complex errors requiring minimal paraphrasing}\\
\texttt{25. UNK: Leave unchanged if error is unclear}\\[1mm]
\texttt{RULES:}\\
\texttt{- Make the SMALLEST possible edit to fix each error}\\
\texttt{- Change only what is grammatically or orthographically wrong}\\
\texttt{- Preserve the original meaning and style}\\
\texttt{- Do NOT improve fluency beyond fixing errors}\\
\texttt{- Do NOT make stylistic changes}\\
\texttt{- If no errors exist, return the original sentence unchanged}\\[1mm]
\texttt{Few-shot examples from BEA train:}\\
\texttt{Follow the same input-to-correction style. Do not copy these examples; use them only as guidance.}\\[1mm]
\texttt{Example 1:}\\
\texttt{Input: Alison put on her coat, close the door and went to his parents house by car.}\\
\texttt{Correction: Alison put on her coat, closed the door and went to her parents ' house by car.}\\[1mm]
\texttt{Example 2:}\\
\texttt{Input: we can see the number of the projected sales of jeans of Jack\&Jones Co. will be at 450 thousands of pairs in next Jan, it will be higher than the number of the projected sales of jeans of Mango Co. about 300 thousands of pairs.Then the number of Jack\&Jones Co. will drop to 250 thousands of pairs in Feb next year.}\\
\texttt{Correction: we can see the number of projected sales of jeans of Jack\&Jones Co. will be at 450 thousand pairs next Jan, it will be higher than the number of the projected sales of jeans of Mango Co.; about 300 thousand pairs. Then the number of Jack\&Jones Co. will drop to 250 thousand pairs in Feb next year.}\\[1mm]
\texttt{Example 3:}\\
\texttt{Input: The girl is 18 years old and men (as it later turns out) are vampires and they have over 100 years.}\\
\texttt{Correction: The girl is 18 years old and the men (as it later turns out) are vampires and they are over 100 years old.}\\[1mm]
\texttt{Example 4:}\\
\texttt{Input: Like what was once, I have no doubt that this world of wonders will remain important in the coming years.}\\
\texttt{Correction: Like what was once, I have no doubt that this world of wonders will remain important in the coming years.}\\[1mm]
\texttt{Example 5:}\\
\texttt{Input: The need of export and import items has increased in the last 10 years consequently the life of the people has raised in faster respond.}\\
\texttt{Correction: The need for exporting and importing items has increased in the last 10 years. Consequently, the life of the people has been raised in rapid response.}\\[1mm]
\texttt{Example 6:}\\
\texttt{Input: In conclusion, it's very important to learn the concept than saying something stupid....}\\
\texttt{Correction: In conclusion, it's more important to learn the concept than say saying something stupid....}\\[1mm]
\texttt{Example 7:}\\
\texttt{Input: Abortion is an immoral act that people make as everyone has a chance to survive and nobody has the right to decide for anyone life.}\\
\texttt{Correction: Abortion is an immoral act that people perform as everyone has the chance to survive and nobody has the right to decide about anyone's life.}\\[1mm]
\texttt{Example 8:}\\
\texttt{Input: Hi John:}\\
\texttt{Correction: Hi John:}
\end{promptbox}

\subsection{Minimal-edits few-shot + taxonomy + optimized}
\label{prompt:apo}

\subsubsection{Minimal-edits few-shot + taxonomy + optimized for Qwen3-8B}
\label{prompt:apo:qwen}

\begin{promptbox}
\small
\texttt{You are a grammatical error correction system. Make MINIMAL, PRECISE edits to fix errors. DO NOT rewrite or paraphrase. Only fix clear grammatical and spelling errors.}\\[1mm]
\texttt{Focus on these 25 error types:}\\[1mm]
\texttt{WORD-LEVEL ERRORS:}\\
\texttt{1. ADJ: Wrong adjective choice (big$\to$wide)}\\
\texttt{2. ADJ:FORM: Adjective form errors - comparatives/superlatives (goodest$\to$best, more easy$\to$easier)}\\
\texttt{3. ADV: Wrong adverb choice (speedily$\to$quickly)}\\
\texttt{4. CONJ: Wrong conjunction (and$\to$but)}\\
\texttt{5. CONTR: Contraction errors (n't$\to$not)}\\
\texttt{6. DET: Wrong/missing/extra determiner (the$\to$a, $\emptyset\to$the, the$\to\emptyset$)}\\
\texttt{7. NOUN: Wrong noun choice (person$\to$people)}\\
\texttt{8. NOUN:INFL: Count-mass noun errors (informations$\to$information)}\\
\texttt{9. NOUN:NUM: Noun number agreement (cat$\to$cats)}\\
\texttt{10. NOUN:POSS: Noun possessive errors (friends$\to$friend's)}\\
\texttt{11. PART: Wrong particle (look in$\to$look at)}\\
\texttt{12. PREP: Wrong/missing/extra preposition (of$\to$at, $\emptyset\to$at, at$\to\emptyset$)}\\
\texttt{13. PRON: Wrong pronoun (ours$\to$ourselves)}\\
\texttt{14. VERB: Wrong verb choice (ambulate$\to$walk)}\\
\texttt{15. VERB:FORM: Verb form errors - infinitive/gerund/participle (to eat$\to$eating, dancing$\to$danced)}\\
\texttt{16. VERB:INFL: Verb inflection errors (getted$\to$got, fliped$\to$flipped)}\\
\texttt{17. VERB:SVA: Subject-verb agreement ((He) have$\to$(He) has)}\\
\texttt{18. VERB:TENSE: Verb tense errors including modals and passive (eats$\to$ate, eats$\to$can eat, eats$\to$was eaten)}\\[1mm]
\texttt{MECHANICAL ERRORS:}\\
\texttt{19. ORTH: Orthography - capitalization/whitespace (Bestfriend$\to$best friend, THIS$\to$this)}\\
\texttt{20. PUNCT: Punctuation errors (!$\to$., missing commas, extra periods)}\\
\texttt{21. SPELL: Spelling errors (genectic$\to$genetic, color$\to$colour)}\\
\texttt{22. WO: Word order errors (only can$\to$can only)}\\[1mm]
\texttt{OTHER:}\\
\texttt{23. MORPH: Morphology - same lemma, different part of speech (quick[adj]$\to$quickly[adv])}\\
\texttt{24. OTHER: Complex errors requiring minimal paraphrasing}\\
\texttt{25. UNK: Leave unchanged if error is unclear}\\[1mm]
\texttt{RULES:}\\
\texttt{- Make the SMALLEST possible edit to fix each error}\\
\texttt{- Change only what is grammatically or orthographically wrong}\\
\texttt{- Preserve the original meaning and style}\\
\texttt{- Do NOT improve fluency beyond fixing errors}\\
\texttt{- If no errors exist, return the original sentence unchanged}\\
\texttt{- Output plain text only: NEVER use Markdown or any markup in the output - no **bold**, no *italics*, no backticks. Return the corrected sentence exactly as plain text}\\
\texttt{- PUNCT: when a sentence starts with an introductory word, phrase or clause, insert the missing comma after it (However->However, | Also->Also, | Nowadays->Nowadays, | Finally, So, Luckily, Unfortunately, Today, Instead, Actually, For example, In my opinion, One day, The next day, Before that, As a rule, To summarise, Once upon a time, and time/place openers like 'About 40 years ago' or 'In a car')}\\[1mm]
\texttt{Correct each numbered text independently. Return only the corrected texts in the same numbered format, one per line.}\\[1mm]
\texttt{\#\#\# Example}\\
\texttt{\#\#\# Texts to correct:}\\
\texttt{Sentence 1: Alison put on her coat, close the door and went to his parents house by car.}\\
\texttt{Sentence 2: we can see the number of the projected sales of jeans of Jack\&Jones Co. will be at 450 thousands of pairs in next Jan, it will be higher than the number of the projected sales of jeans of Mango Co. about 300 thousands of pairs.Then the number of Jack\&Jones Co. will drop to 250 thousands of pairs in Feb next year.}\\
\texttt{Sentence 3: The girl is 18 years old and men (as it later turns out) are vampires and they have over 100 years.}\\
\texttt{Sentence 4: Like what was once, I have no doubt that this world of wonders will remain important in the coming years.}\\
\texttt{Sentence 5: The need of export and import items has increased in the last 10 years consequently the life of the people has raised in faster respond.}\\
\texttt{Sentence 6: In conclusion, it's very important to learn the concept than saying something stupid....}\\
\texttt{Sentence 7: Abortion is an immoral act that people make as everyone has a chance to survive and nobody has the right to decide for anyone life.}\\
\texttt{Sentence 8: Hi John:}\\[1mm]
\texttt{\#\#\# Corrected texts:}\\
\texttt{Sentence 1: Alison put on her coat, closed the door and went to her parents ' house by car.}\\
\texttt{Sentence 2: we can see the number of projected sales of jeans of Jack\&Jones Co. will be at 450 thousand pairs next Jan, it will be higher than the number of the projected sales of jeans of Mango Co.; about 300 thousand pairs. Then the number of Jack\&Jones Co. will drop to 250 thousand pairs in Feb next year.}\\
\texttt{Sentence 3: The girl is 18 years old and the men (as it later turns out) are vampires and they are over 100 years old.}\\
\texttt{Sentence 4: Like what was once, I have no doubt that this world of wonders will remain important in the coming years.}\\
\texttt{Sentence 5: The need for exporting and importing items has increased in the last 10 years. Consequently, the life of the people has been raised in rapid response.}\\
\texttt{Sentence 6: In conclusion, it's more important to learn the concept than say saying something stupid....}\\
\texttt{Sentence 7: Abortion is an immoral act that people perform as everyone has the chance to survive and nobody has the right to decide about anyone's life.}\\
\texttt{Sentence 8: Hi John:}\\[1mm]
\texttt{\#\#\# Texts to correct:}\\
\texttt{\{\}}\\[1mm]
\texttt{\#\#\# Corrected texts:}
\end{promptbox}

\subsubsection{Minimal-edits few-shot + taxonomy + optimized for GPT-4.1-mini}
\label{prompt:apo:gpt}

\begin{promptbox}
\small
\texttt{You are a grammatical error correction system. Make MINIMAL, PRECISE edits to fix errors.}\\
\texttt{DO NOT rewrite or paraphrase. Only fix clear grammatical and spelling errors.}\\[1mm]
\texttt{RULES:}\\
\texttt{- Make the SMALLEST possible edit to fix each error}\\
\texttt{- Change only what is grammatically or orthographically wrong}\\
\texttt{- Preserve the original meaning and style}\\
\texttt{- Do NOT improve fluency beyond fixing errors}\\
\texttt{- Do NOT make stylistic changes}\\
\texttt{- Comma-splice fix: when a comma joins TWO COMPLETE INDEPENDENT CLAUSES (each with its own subject + finite}\\
\texttt{verb), replace the comma with a period and capitalize the next word. ONLY apply when the second clause clearly}\\
\texttt{starts with a subject pronoun (I, We, They, He, She, It) followed by its own verb.}\\
\texttt{~~~~`I was tired , I went home' $\to$ `I was tired . I went home'}\\
\texttt{~~~~`We waited , they arrived late' $\to$ `We waited . They arrived late'}\\
\texttt{~~~~`I followed them , I learned a lot' $\to$ `I followed them . I learned a lot'}\\
\texttt{~~~~Do NOT apply when the second part is a dependent clause, fragment, or list continuation. When in doubt,}\\
\texttt{~~~~leave the comma alone.}\\
\texttt{- Past-narrative tense consistency: when a clause is clearly set in the past (signaled by an earlier finite}\\
\texttt{past-tense verb in the same sentence --- was, were, did, said, went, came, told, started, entered, etc., OR}\\
\texttt{a past-time adverbial --- `yesterday', `last <time>', `<N> years ago', `when I was <X>'), AND a later verb}\\
\texttt{in the same sentence is in present tense referring to the SAME past event, change that verb to its}\\
\texttt{simple-past form.}\\
\texttt{~~~~`After many years he still dream to become a hero' $\to$ `... still dreamt ...'}\\
\texttt{~~~~`He entered the university because he thinks it is good' $\to$ `... thought it was good'}\\
\texttt{~~~~`When I was a child , I play with toys' $\to$ `... I played with toys'}\\
\texttt{~~~~Do NOT apply when the present-tense verb expresses a general truth interjected in the past narrative}\\
\texttt{~~~~(`I learned that water boils at 100 degrees'). When in doubt, leave the verb alone.}\\
\texttt{- If no errors exist, return the original sentence unchanged}\\[1mm]
\texttt{Few-shot examples from BEA train:}\\
\texttt{Follow the same input-to-correction style. Do not copy these examples; use them only as guidance.}\\[1mm]
\texttt{Example 1:}\\
\texttt{Input: Yours sincerely}\\
\texttt{Correction: Yours sincerely}\\[1mm]
\texttt{Example 2:}\\
\texttt{Input: I'd love to hear from you and maybe we can make some plans to meet up.}\\
\texttt{Correction: I'd love to hear from you and maybe we can make some plans to meet up.}\\[1mm]
\texttt{Example 3:}\\
\texttt{Input: Although the maintenance cost of car is quite expensive, it does not stop people from using cars since convenience matters to them the most and not money.}\\
\texttt{Correction: Although the maintenance cost of car is quite expensive, it does not stop people from using cars, since convenience matters to them the most and not money.}\\[1mm]
\texttt{Example 4:}\\
\texttt{Input: I studied in Kuwait schools, after that I joined Damascus university, the faculty of human medicine, I graduated from it in 2008 with very good grade, then I got the Master's degree in laboratory medicine from the same university (Ministry of Higher Education) in 2013 with Excellent grade.}\\
\texttt{Correction: I studied in Kuwaiti schools. After that I went to Damascus university, the faculty of human medicine. I graduated from it in 2008 with a very good grade, then I got the Master's degree in laboratory medicine from the same university (Ministry of Higher Education) in 2013 with an Excellent grade.}\\[1mm]
\texttt{Example 5:}\\
\texttt{Input: The female respondents are more fond of reading books that the male ones.}\\
\texttt{Correction: The female respondents are more fond of reading books than the male ones.}\\[1mm]
\texttt{Example 6:}\\
\texttt{Input: I was gardening since the beginning of the morning, when the postman arrived.}\\
\texttt{Correction: I had been gardening since the beginning of the morning, when the postman arrived.}\\[1mm]
\texttt{Example 7:}\\
\texttt{Input: And you enjoy between your mates and live all those special moments with them.}\\
\texttt{Correction: And you enjoy with between your mates and experience all those special moments with them.}\\[1mm]
\texttt{Example 8:}\\
\texttt{Input: The Vietnamese education and study system is completely different from the Australian one.}\\
\texttt{Correction: The Vietnamese education and study system is completely different from the Australian one.}
\end{promptbox}

\subsubsection{Minimal-edits few-shot + taxonomy + optimized for Claude Sonnet~4.6}
\label{prompt:apo:claude}

\begin{promptbox}
\small
\texttt{You are a grammatical error correction system. Make MINIMAL, PRECISE edits to fix errors.}\\
\texttt{DO NOT rewrite or paraphrase. Only fix clear grammatical and spelling errors.}\\[1mm]
\texttt{Focus on these 25 error types:}\\[1mm]
\texttt{WORD-LEVEL ERRORS:}\\
\texttt{1. ADJ: Wrong adjective choice (big$\to$wide)}\\
\texttt{2. ADJ:FORM: Adjective form errors - comparatives/superlatives (goodest$\to$best,}\\
\texttt{more easy$\to$easier)}\\
\texttt{3. ADV: Wrong adverb choice (speedily$\to$quickly)}\\
\texttt{4. CONJ: Wrong conjunction (and$\to$but)}\\
\texttt{5. CONTR: Contraction errors (n't$\to$not)}\\
\texttt{6. DET: Wrong/missing/extra determiner (the$\to$a, $\emptyset\to$the, the$\to\emptyset$)}\\
\texttt{7. NOUN: Wrong noun choice (person$\to$people)}\\
\texttt{8. NOUN:INFL: Count-mass noun errors (informations$\to$information)}\\
\texttt{9. NOUN:NUM: Noun number agreement (cat$\to$cats)}\\
\texttt{10. NOUN:POSS: Noun possessive errors (friends$\to$friend's)}\\
\texttt{11. PART: Wrong particle (look in$\to$look at)}\\
\texttt{12. PREP: Wrong/missing/extra preposition (of$\to$at, $\emptyset\to$at, at$\to\emptyset$)}\\
\texttt{13. PRON: Wrong pronoun (ours$\to$ourselves)}\\
\texttt{14. VERB: Wrong verb choice (ambulate$\to$walk)}\\
\texttt{15. VERB:FORM: Verb form errors - infinitive/gerund/participle (to eat$\to$eating,}\\
\texttt{dancing$\to$danced)}\\
\texttt{16. VERB:INFL: Verb inflection errors (getted$\to$got, fliped$\to$flipped)}\\
\texttt{17. VERB:SVA: Subject-verb agreement ((He) have$\to$(He) has)}\\
\texttt{18. VERB:TENSE: Verb tense errors including modals and passive (eats$\to$ate, eats$\to$can eat,}\\
\texttt{eats$\to$was eaten)}\\[1mm]
\texttt{MECHANICAL ERRORS:}\\
\texttt{19. ORTH: Orthography - capitalization/whitespace (Bestfriend$\to$best friend, THIS$\to$this)}\\
\texttt{20. PUNCT: Punctuation errors (!$\to$., missing commas, extra periods)}\\
\texttt{21. SPELL: Spelling errors (genectic$\to$genetic, color$\to$colour)}\\
\texttt{22. WO: Word order errors (only can$\to$can only)}\\[1mm]
\texttt{OTHER:}\\
\texttt{23. MORPH: Morphology - same lemma, different part of speech (quick[adj]$\to$quickly[adv])}\\
\texttt{24. OTHER: Complex errors requiring minimal paraphrasing}\\
\texttt{25. UNK: Leave unchanged if error is unclear}\\[1mm]
\texttt{RULES:}\\
\texttt{- Make the SMALLEST possible edit to fix each error}\\
\texttt{- Change only what is grammatically or orthographically wrong}\\
\texttt{- Preserve the original meaning and style}\\
\texttt{- Do NOT improve fluency beyond fixing errors}\\
\texttt{- Do NOT make stylistic changes}\\
\texttt{- If no errors exist, return the original sentence unchanged}\\[1mm]
\texttt{PUNCTUATION --- sentence-initial intro comma (apply when missing):}\\
\texttt{When a sentence begins with an introductory dependent clause or adverbial phrase, and that intro is followed}\\
\texttt{directly by the main clause (subject + finite verb), insert a comma between the intro and the main clause.}\\
\texttt{Triggers at the sentence start:}\\
\texttt{~~(a) Subordinator clause: When / If / While / Although / Because / Since / After / Before / Unless / Until /}\\
\texttt{~~As / Whenever <clause> <subject> <verb>.}\\
\texttt{~~(b) Adverbial connector or phrase: However / Also / Therefore / Thus / Moreover / Furthermore / Indeed /}\\
\texttt{~~Nowadays / Generally / Personally / Of course / As a result / On the other hand / On the one hand /}\\
\texttt{~~For example / For instance / In addition / In my opinion / In my experience / In conclusion /}\\
\texttt{~~First of all / Firstly / Secondly / Finally.}\\
\texttt{Examples:}\\
\texttt{~~`When I was younger I used to say...' $\to$ `When I was younger , I used to say...'}\\
\texttt{~~`In my experience when I do n't have a car...' $\to$ `In my experience , when I do n't have a car...'}\\
\texttt{~~`For example Barcelona , Madrid , Sevilla' $\to$ `For example , Barcelona , Madrid , Sevilla'}\\
\texttt{~~`However the weather was bad' $\to$ `However , the weather was bad'}\\
\texttt{Constraints:}\\
\texttt{~~- Apply ONLY at the sentence-initial position (token 1 of a sentence, or directly after a sentence-final}\\
\texttt{~~period that starts a new sentence).}\\
\texttt{~~- Do NOT add commas mid-sentence, before `and' / `but' / `or', or for stylistic flow.}\\
\texttt{~~- Do NOT remove existing commas.}\\
\texttt{~~- If the comma is already present after the intro, leave the sentence unchanged.}\\[1mm]
\texttt{PUNCTUATION --- comma-splice fix (replace `,' with `. <Capital>'):}\\
\texttt{When a comma joins TWO COMPLETE INDEPENDENT CLAUSES (each with its own subject + finite verb), replace the}\\
\texttt{comma with a period and capitalize the next word.}\\
\texttt{ONLY apply when the second clause clearly starts with one of these subject pronouns / demonstratives followed}\\
\texttt{by its own finite verb: I / We / You / They / He / She / It / This / That / These / Those.}\\
\texttt{Examples:}\\
\texttt{~~`I was tired , I went home' $\to$ `I was tired . I went home'}\\
\texttt{~~`We waited , they arrived late' $\to$ `We waited . They arrived late'}\\
\texttt{~~`Thank you for your e-mail , it was wonderful' $\to$ `Thank you for your e-mail . It was wonderful'}\\
\texttt{~~`Public transport is important , it brings benefits' $\to$ `Public transport is important . It brings benefits'}\\
\texttt{~~`He is the best player , that is why he won' $\to$ `He is the best player . That is why he won'}\\
\texttt{Do NOT apply when:}\\
\texttt{~~- The second part is a dependent clause, fragment, or list continuation.}\\
\texttt{~~- The second part starts with a lowercase word, a non-pronoun noun, or a non-finite verb form.}\\
\texttt{~~- The comma is already correct (e.g., between list items, between adjectives).}\\
\texttt{When in doubt, leave the comma alone.}\\[1mm]
\texttt{Few-shot examples from BEA train:}\\
\texttt{Follow the same input-to-correction style. Do not copy these examples; use them only as guidance.}\\[1mm]
\texttt{Example 1:}\\
\texttt{Input: Alison put on her coat, close the door and went to his parents house by car.}\\
\texttt{Correction: Alison put on her coat, closed the door and went to her parents ' house by car.}\\[1mm]
\texttt{Example 2:}\\
\texttt{Input: we can see the number of the projected sales of jeans of Jack\&Jones Co. will be at 450 thousands of pairs in next Jan, it will be higher than the number of the projected sales of jeans of Mango Co. about 300 thousands of pairs.Then the number of Jack\&Jones Co. will drop to 250 thousands of pairs in Feb next year.}\\
\texttt{Correction: we can see the number of projected sales of jeans of Jack\&Jones Co. will be at 450 thousand pairs next Jan, it will be higher than the number of the projected sales of jeans of Mango Co.; about 300 thousand pairs. Then the number of Jack\&Jones Co. will drop to 250 thousand pairs in Feb next year.}\\[1mm]
\texttt{Example 3:}\\
\texttt{Input: The girl is 18 years old and men (as it later turns out) are vampires and they have over 100 years.}\\
\texttt{Correction: The girl is 18 years old and the men (as it later turns out) are vampires and they are over 100 years old.}\\[1mm]
\texttt{Example 4:}\\
\texttt{Input: Like what was once, I have no doubt that this world of wonders will remain important in the coming years.}\\
\texttt{Correction: Like what was once, I have no doubt that this world of wonders will remain important in the coming years.}\\[1mm]
\texttt{Example 5:}\\
\texttt{Input: The need of export and import items has increased in the last 10 years consequently the life of the people has raised in faster respond.}\\
\texttt{Correction: The need for exporting and importing items has increased in the last 10 years. Consequently, the life of the people has been raised in rapid response.}\\[1mm]
\texttt{Example 6:}\\
\texttt{Input: In conclusion, it's very important to learn the concept than saying something stupid....}\\
\texttt{Correction: In conclusion, it's more important to learn the concept than say saying something stupid....}\\[1mm]
\texttt{Example 7:}\\
\texttt{Input: Abortion is an immoral act that people make as everyone has a chance to survive and nobody has the right to decide for anyone life.}\\
\texttt{Correction: Abortion is an immoral act that people perform as everyone has the chance to survive and nobody has the right to decide about anyone's life.}\\[1mm]
\texttt{Example 8:}\\
\texttt{Input: Hi John:}\\
\texttt{Correction: Hi John:}
\end{promptbox}

\subsubsection{Minimal-edits few-shot + taxonomy + optimized for Gemini~3-Flash}
\label{prompt:apo:gemini}

\begin{promptbox}
\small
\texttt{You are a precision-focused grammatical error correction system.}\\[1mm]
\texttt{CORE PRINCIPLE: Only fix CLEAR, UNAMBIGUOUS errors. When in doubt, leave the original unchanged.}\\
\texttt{Every unnecessary edit is a mistake.}\\[1mm]
\texttt{FIX:}\\
\texttt{- Spelling errors (genectic$\to$genetic)}\\
\texttt{- Subject-verb agreement (He have$\to$He has)}\\
\texttt{- Verb form: tense, infinitive/gerund, inflection (I looking$\to$I am looking, getted$\to$got)}\\
\texttt{- Missing/wrong/extra determiners (I have cat$\to$I have a cat)}\\
\texttt{- Missing/wrong/extra prepositions (interested at$\to$interested in)}\\
\texttt{- Noun number (many cat$\to$many cats, informations$\to$information)}\\
\texttt{- Possessive errors (friends book$\to$friend's book)}\\
\texttt{- Pronoun errors (me went$\to$I went)}\\
\texttt{- Adjective/adverb form (more easy$\to$easier, quick$\to$quickly)}\\
\texttt{- Capitalization of common nouns (Public transport$\to$public transport) and sentence-initial (i$\to$I)}\\
\texttt{- Word order when grammatically wrong (I always am$\to$I am always)}\\
\texttt{- Missing words needed for grammar (I want go$\to$I want to go)}\\
\texttt{- ONLY these punctuation fixes: add comma after introductory adverbs/transitions (However/Also/For example/}\\
\texttt{First of all + comma), fix double periods (..), fix missing period at sentence end. Do NOT add commas}\\
\texttt{elsewhere. Do NOT remove existing commas.}\\[1mm]
\texttt{DO NOT:}\\
\texttt{- Replace words with synonyms (big$\to$large, start$\to$begin)}\\
\texttt{- Rewrite or restructure sentences}\\
\texttt{- Add commas for style or readability}\\
\texttt{- Change style or register}\\[1mm]
\texttt{Examples:}\\
\texttt{Input: I want go to the store and buy some foods .}\\
\texttt{Output: I want to go to the store and buy some food .}\\[1mm]
\texttt{Input: She do n't likes swimming but her brother do .}\\
\texttt{Output: She does n't like swimming but her brother does .}\\[1mm]
\texttt{Input: In my opinion the Public transport is very important for citys .}\\
\texttt{Output: In my opinion , public transport is very important for cities .}\\[1mm]
\texttt{Input: Yesterday I have went to the library and readed many book .}\\
\texttt{Output: Yesterday I went to the library and read many books .}\\[1mm]
\texttt{Input: Maybe I will change my mind , maybe not .}\\
\texttt{Output: Maybe I will change my mind , maybe not .}\\[1mm]
\texttt{If the sentence has no clear errors, return it EXACTLY as given.}
\end{promptbox}

\clearpage
\section{Inference Pipeline and Structured Output}
\label{app:inference-pipeline}

Each input sentence is processed independently through a single LLM call, with no cross-sentence batching in
the single-pass setting (the batching variants of Section~\ref{sec:experimental_setup} simply pack multiple
sentences into one call before invoking the same pipeline).

The agent issues one chat completion through a LiteLLM router that abstracts over the underlying provider
(OpenAI, Anthropic, Google). The system message is the configured prompt template; the user message is the
raw source sentence. To eliminate free-form post-processing of model output, we constrain the response with a
JSON schema derived from a Pydantic model and attached to the request as a strict
\texttt{response\_format}. Field descriptions declared on the Pydantic model propagate into the schema and
act as in-band instructions to the model.

\paragraph{Schema definition.}
The response contract is declared once as a Pydantic class:

\begin{lstlisting}[language=Python,basicstyle=\ttfamily\footnotesize,frame=single,breaklines=true,
  columns=fullflexible,breakatwhitespace=true,
  postbreak=\mbox{\textcolor{red}{$\hookrightarrow$}\space}]
class GECResponse(BaseModel):
    corrected_sentence: str = Field(
        ..., description="Corrected version of the input sentence")
\end{lstlisting}

\paragraph{Generated JSON schema.}
At call time, the class is converted to JSON Schema, all object nodes are closed with
\texttt{additionalProperties: false}, and the result is wrapped into the provider-agnostic
\texttt{response\_format} envelope:

\begin{lstlisting}[basicstyle=\ttfamily\footnotesize,frame=single,breaklines=true,
  columns=fullflexible,breakatwhitespace=true,
  postbreak=\mbox{\textcolor{red}{$\hookrightarrow$}\space}]
{
  "type": "json_schema",
  "json_schema": {
    "name": "GECResponse",
    "strict": true,
    "schema": {
      "type": "object",
      "additionalProperties": false,
      "required": ["corrected_sentence"],
      "properties": {
        "corrected_sentence": {
          "type": "string",
          "description": "Corrected version of the input sentence"
        }
      }
    }
  }
}
\end{lstlisting}

\begin{figure}[t]
\centering
\small
\begin{tikzpicture}[
    node distance=5mm,
    every node/.style={font=\footnotesize},
    box/.style={rectangle, draw, rounded corners=2pt,
                align=center, inner sep=3pt, minimum width=42mm},
    arrow/.style={-{Latex[length=1.6mm]}, semithick},
]
\node[box]                       (src)    {source sentence};
\node[box, below=of src]         (call)   {\texttt{router.completion(\dots)}\\
                                          structured \texttt{response\_format}};
\node[box, below=of call]        (router) {LiteLLM Router\\
                                          $\rightarrow$ provider};
\node[box, below=of router]      (val)    {\texttt{GECResponse.model\_validate}};
\node[box, below=of val]         (out)    {\texttt{predictions.eval.txt}};
\draw[arrow] (src)    -- (call);
\draw[arrow] (call)   -- (router);
\draw[arrow] (router) -- (val);
\draw[arrow] (val)    -- (out);
\end{tikzpicture}

\vspace{2mm}

\begin{tabular}{@{}ll@{}}
\toprule
\textbf{Parameter} & \textbf{Value (from run YAML)} \\
\midrule
\texttt{model}             & \texttt{gpt-4.1-mini} \\
\texttt{temperature}       & \texttt{0.0} \\
\texttt{top\_p}            & \texttt{0.1} \\
\texttt{reasoning\_effort} & \texttt{None} \\
\texttt{timeout}           & \texttt{90} s \\
\texttt{response\_format}  & \texttt{json\_schema(GECResponse)} \\
\bottomrule
\end{tabular}

\caption{Per-sentence inference flow. Every sentence goes through one structured-output call. Decoding
parameters are read verbatim from the run YAML, ensuring deterministic replay.}
\label{fig:inference-flow}
\end{figure}

The decoding parameters in Figure~\ref{fig:inference-flow} are read from the run YAML and the exact
configuration file is copied into the output directory alongside \texttt{results.json}, so a run can be
replayed bit-for-bit given the same provider model snapshot. The returned payload is validated with
\texttt{model\_validate}, so any schema violation is caught deterministically rather than being masked by
string heuristics. If a provider rejects \texttt{json\_schema}, the client transparently retries with
\texttt{response\_format = \{"type":"json\_object"\}}; for the single-field case, a final recovery path
extracts the corrected sentence from malformed JSON to keep evaluation aligned. This design ensures that
every sentence yields exactly one validated correction, making the sentence-to-prediction mapping bijective
and the run reproducible given a fixed configuration.

\paragraph{Batch user payload.}
  The user message is a single deterministic string in which each
  source carries an explicit 1-based identifier; this gives the model a
  stable handle for cross-referencing in the response:
\begin{tcolorbox}[
    enhanced, colback=gray!5, colframe=black!60,
    boxrule=0.4pt, arc=1pt,
    left=6pt, right=6pt, top=4pt, bottom=4pt,
    fontupper=\ttfamily\small,
    width=\linewidth
  ]
  Correct each sentence separately. Return a result for each SENTENCE\_ID. \\[2pt]
  Sentence 1: \textless source 1\textgreater \\
  Sentence 2: \textless source 2\textgreater \\
  \dots \\
  Sentence k: \textless source k\textgreater
  \end{tcolorbox}

  \paragraph{Batch response schema.}
  The response contract is again a Pydantic model, so the same strict
  \texttt{response\_format=json\_schema(\dots)} mechanism applies.
  The schema enforces that the model returns exactly one
  \texttt{CORRECTED\_SENTENCE} per \texttt{SENTENCE\_ID}:

\begin{lstlisting}[
    language=Python,
    basicstyle=\ttfamily\footnotesize,
    keywordstyle=\color{blue!70!black}\bfseries,
    stringstyle=\color{green!45!black},
    commentstyle=\color{gray}\itshape,
    showstringspaces=false,
    columns=fullflexible,
    keepspaces=true,
    frame=single,
    framesep=6pt,
    xleftmargin=10pt,
    breaklines=true,
    breakatwhitespace=true,
    postbreak=\mbox{\textcolor{gray}{$\hookrightarrow$}\space},
    lineskip=1pt,
]
class BatchSentenceGECItem(BaseModel):
    SENTENCE_ID: int = Field(
        ..., description="1-based index within the batch input."
    )
    CORRECTED_SENTENCE: str = Field(
        ..., description="Corrected version of the sentence."
    )

class BatchGECResponse(BaseModel):
    SENTENCES: list[BatchSentenceGECItem] = Field(
        ..., description="Corrections for each input sentence."
    )
\end{lstlisting}

\clearpage
\section{LLM-assisted Prompt Optimization Skill}
\label{app:apo-skill}

The LLM-assisted Prompt Optimization loop described in Section~\ref{sec:apo} is implemented as a Claude Code skill
(\texttt{gec-prompt-optimise}) that the optimizer agent loads at run-time. The skill is a single Markdown file
with YAML front-matter; the agent treats it as an instruction sheet that defines the
diagnose-propose-register-eval-log cycle. We reproduce its full content below as a single User-Prompt block, the
form in which the optimizer agent receives it.

\begin{promptbox}
\begin{lstlisting}[style=mdcode]
---
name: gec-prompt-optimise
description: Propose the next GEC system-prompt iteration by running an
  ERRANT per-category diagnostic on the current best run, picking ONE
  targeted rule, and registering a new prompt_NN_*.py + matching config
  + RESULTS.md row. Use when the user asks to iterate the prompt,
  propose a next variant, beat the current best, or "what should we try
  next" for BEA/CoNLL GEC.
---
\end{lstlisting}

\paragraph{GEC prompt optimisation, rule-driven iteration.}

You iterate prompts by \textbf{diagnose $\to$ propose-one-change $\to$ register $\to$ eval $\to$ log}. Every
iteration changes exactly one variable so deltas are attributable. No automated search (no GEPA, no APO), this
skill is purely rule-driven.

The canonical example to mimic is
\texttt{src/agents/prompts/\allowbreak prompt\_\allowbreak 19\_\allowbreak taxonomy\_\allowbreak optimised\_\allowbreak
gemini\_\allowbreak family\_\allowbreak v5\_\allowbreak orth\_\allowbreak tight\_\allowbreak punct\_\allowbreak
explicit\_\allowbreak comma\_\allowbreak splice.py}, read its header before you start.

\paragraph{Phase 1: Diagnose the current best.}

\begin{enumerate}
\item \textbf{Find the parent.} Ask the user which prompt to iterate on, or scan \texttt{RESULTS.md} for the
\textbf{bolded} best-$F_{0.5}$ row under the target model section. Identify the matching
\texttt{outputs/<run>/} directory by the
\texttt{<split>\allowbreak\_<model>\allowbreak\_<prompt\_name>\allowbreak\_...\_\allowbreak batch\_<N>\allowbreak\_temp...} naming convention.

\item \textbf{Get per-ERRANT-category P/R/$F_{0.5}$.} The codebase only writes overall metrics; the category
breakdown has to be regenerated. Run:

\begin{lstlisting}[style=mdcode,language=bash]
PARENT=outputs/<parent-run>
# or conll-2014 / test src, match run
SPLIT_SRC=data/en_bea/dev/bea-dev.src
SPLIT_M2=data/en_bea/dev/bea-dev.m2

uv run errant_parallel \
    -orig "$SPLIT_SRC" \
    -cor \
    "$PARENT/predictions.eval.txt" \
    -out "$PARENT/predictions.m2"

uv run errant_compare \
    -hyp "$PARENT/predictions.m2" \
    -ref "$SPLIT_M2" -cat 2 \
    | tee \
    "$PARENT/errant_per_category.txt"
\end{lstlisting}

\texttt{-cat 2} is type-level (PUNCT, DET, ORTH, SPELL, MORPH, VERB:TENSE, NOUN:NUM, ...). Do not use
\texttt{-cat 1} (operation only) or \texttt{-cat 3} (op+type), type-level is what surfaces actionable rules.

\item \textbf{Rank categories.} Pull the table from \texttt{errant\_per\_category.txt}. Sort by FN descending
for ``missed corrections that we could add rules for''; sort by FP descending for ``over-corrections that need
a tightening rule''. The biggest $F_{0.5}$ movers are usually PUNCT, DET, ORTH, SPELL, MORPH, VERB:TENSE,
NOUN:NUM.

\item \textbf{Inspect 10--20 real examples in the target category.} Don't propose a rule blind. Find sentences
where the reference makes that edit and the prediction doesn't (or makes a wrong one). Quick recipe:

\begin{lstlisting}[style=mdcode,language=bash]
# diff prediction vs reference line-by-line;
# line numbers map to .src/.ref0/.m2
paste -d'\t' \
    "$SPLIT_SRC" \
    data/en_bea/dev/bea-dev.ref0 \
    "$PARENT/predictions.eval.txt" \
    | awk -F'\t' '$2 != $3' | head -40
\end{lstlisting}

Then read \texttt{\$SPLIT\_M2} around those sentence indices to confirm the error-type label.

\item \textbf{Write the diagnostic in one sentence:} \texttt{"<CAT> had <FN> FN. Of those, \textasciitilde<estimate>
are <specific targetable sub-pattern>."} If you can't write a \emph{specific sub-pattern}, the category isn't
ready for a rule yet, pick the next-worst category.
\end{enumerate}

\paragraph{Phase 2: Propose exactly one change.}

\textbf{Rules for the proposal, non-negotiable:}

\begin{itemize}
\item \textbf{One variable per iteration.} Either \emph{add one narrow targeted rule} OR \emph{tighten one
existing rule}, never both, never two new rules, never a rule plus a reorder. If you have two good ideas, split
into two iterations.
  \begin{itemize}
  \item \textbf{Why:} the whole point of this loop is attributable deltas. Two changes at once means the next
  iteration's delta is uninterpretable.
  \end{itemize}
\item \textbf{High-precision additions only.} $F_{0.5}$ weights precision 2x recall. A rule that adds 50 TPs and
50 FPs is a wash; 30 TPs / 5 FPs is a win. If the rule can't beat \textasciitilde 80\% precision on the diagnostic
sample you inspected, don't propose it.
\item \textbf{Be specific to the point of pedantry.} ``Fix punctuation'' is not a rule. The rule must be sharp
enough that a junior annotator could apply it consistently. See \texttt{prompt\_19} line 27--40 for the gold
standard.
\item \textbf{Predict the effect explicitly.} State expected TP gain in target category, FP risk in adjacent
categories, and what observation would falsify the rule (e.g.\ ``if PUNCT FP increases by $>$20 we over-fired'').
\item \textbf{No restructure + content change in the same iteration.} If you want to reorganise the prompt's
bullet order or rewrite preamble, do that as its own no-content-change iteration so the reorder cost is
attributable.
\end{itemize}

\paragraph{Phase 3: Register the new prompt.}

Pick \texttt{NN} = next free integer after the highest \texttt{prompt\_NN\_*.py} in
\texttt{src/agents/prompts/}. Three coordinated edits are required, none is optional:

\paragraph{3a. Create \texttt{src/agents/\allowbreak prompts/\allowbreak prompt\_NN\_\allowbreak <parent\_slug>\_\allowbreak <change\_slug>.py}.}

Header comment must follow the lineage convention from \texttt{prompt\_19}:

\begin{lstlisting}[style=mdcode,language=Python]
# vN = vN-1 (<parent prompt_name key>,
#       F0.5=<parent F0.5 to 4dp>) +
#       <one-line change>.
# Diagnostic: <CAT> had <FN> FN. Of
#       those, ~<count> are <specific
#       pattern>.
# Expected: +<N> TPs in <CAT>, minimal
#       FP risk in <other CATs>.
PROMPT_VARIABLE_NAME = (
    "You are a precision-focused "
    "grammatical error correction "
    "system.\n\n"
    ...
)
\end{lstlisting}

Use the same string-concatenation style as the existing prompts. Do not switch to triple-quoted strings,
f-strings, or Jinja, consistency matters for diffs.

\paragraph{3b. Wire into \texttt{src/agents/prompts/base.py}.}

Two edits in this file:
\begin{itemize}
\item Add \texttt{from .prompt\_NN\_<slug> import (PROMPT\_VARIABLE\_NAME)} at the bottom of the import block
(chronological order, not alphabetical, match the existing pattern).
\item Add \texttt{"<prompt\_name\_key>": PROMPT\_VARIABLE\_NAME,} to the \texttt{GEC\_PROMPTS} dict, in the
same group as its parent (gepa group, gemini-tuned group, gpt41mini-tuned group, taxonomy-optimised-gemini-family
group, etc., see the blank-line groupings in the existing dict).
\end{itemize}

The \texttt{<prompt\_name\_key>} is what goes into the config's \texttt{prompt\_name:} field. Keep it short, drop
the \texttt{prompt\_NN\_} and the file's \texttt{\_v<N>} redundancy isn't needed if the key already carries lineage
(\texttt{taxonomy\_optimised\_gemini\_family\_\allowbreak v6\_\allowbreak <change>}).

\paragraph{3c. Create \texttt{config.en.<split>.<short\_change>.yaml}.}

Copy the parent's config (the one in \texttt{outputs/<parent-run>/config.en.*.yaml}) and change \textbf{only}
\texttt{prompt\_name:}. Do not adjust model, batch\_size, temperature, few-shot config, seed, num\_threads,
anything else changing breaks attribution.

For dev iterations name it \texttt{config.en.dev.<short\_change>.yaml}. For test runs, only create the test
config \emph{after} the dev iteration wins.

\paragraph{Phase 4: Eval + log.}

\begin{enumerate}
\item \textbf{Run dev eval:}
\begin{lstlisting}[style=mdcode,language=bash]
uv run python en_main.py \
    --config \
    config.en.dev.<short_change>.yaml
\end{lstlisting}

\item \textbf{Read the new \texttt{outputs/<run>/metrics.txt}.} Compare to parent.

\item \textbf{Re-run the per-category breakdown on the new run} (same commands as Phase 1 step 2) and check that
the targeted category actually moved. If $F_{0.5}$ went up but the targeted category didn't move, you got lucky
elsewhere, flag this; the rule may not be the cause.

\item \textbf{Append a result row to \texttt{RESULTS.md}} under the appropriate model section. Match the
existing column layout exactly (Prompt / Batch / Exact \% / Precision / Recall / $F_{0.5}$). Group with the
parent under the same separator line. Bold the $F_{0.5}$ if it took the lead for that model.

\item \textbf{Append a result note to the prompt file's header:}
\begin{lstlisting}[style=mdcode,language=Python]
# Result: F0.5=<x.xxxx>
#   (Delta=<+/-y.yyyy> vs parent).
#   Target CAT: <CAT>
#   F0.5 <before>-><after>.
\end{lstlisting}
This closes the loop, next iteration's parent inspection can read the result without re-running anything.

\item \textbf{Decide on the next parent:}
\begin{itemize}
\item \textbf{$F_{0.5}$ improved AND target category moved in the predicted direction:} new prompt is the
parent.
\item \textbf{$F_{0.5}$ improved but target category didn't move:} keep current parent; the win may be noise.
Re-run with a different seed before adopting.
\item \textbf{$F_{0.5}$ unchanged or dropped:} parent stays parent. Next iteration must target a
\emph{different} variable, do not tweak the same rule, that's chasing noise.
\end{itemize}
\end{enumerate}

\paragraph{Hard ``do not'' list.}

\begin{itemize}
\item Do \textbf{not} combine two rules in one iteration, even if both look promising.
\item Do \textbf{not} propose a rule without inspecting actual error examples in Phase 1 step 4.
\item Do \textbf{not} skip the per-category breakdown, overall $F_{0.5}$ hides where the change is actually
landing.
\item Do \textbf{not} modify the agent code (\texttt{src/agents/gec\_agent.py} etc.), the evaluation pipeline,
or unrelated prompts as part of an iteration. Prompt-only changes.
\item Do \textbf{not} change config knobs other than \texttt{prompt\_name:} in the new config, you'd be
confounding the experiment.
\item Do \textbf{not} run GEPA, APO, or any automated search inside this skill. If the user wants those, that's
a separate tool (\texttt{gepa\_optimize.py}); offer it but don't invoke it here.
\item Do \textbf{not} delete or rename the parent prompt file or its config, even if the new prompt wins. Keep
lineage walkable.
\end{itemize}
\end{promptbox}

\clearpage
\section{Batching Sweep and Edit-Count Distributions: Numerical Results}
\label{app:batching-results}
\label{app:edit-distribution}

Table~\ref{tab:batching-best-prompt} reports the batching sweep (batches $B \in \{1, 2, 5, 15, 30,
60, 120\}$) across all seven models under the best-performing manual prompt, minimal-edits
few-shot + taxonomy (Appendix~\ref{prompt:taxonomy-few-shot}). For each (model, batch size $B$)
configuration we report precision, recall, ERRANT $F_{0.5}$, word edit distance (WED), and
Scribendi score.

\begin{table*}[!htbp]
\centering
\small
\setlength{\tabcolsep}{4pt}
\begin{tabular}{l l rrrrrrr}
\toprule
\textbf{Model} & \textbf{Metric} & $\mathbf{B{=}1}$ & $\mathbf{B{=}2}$ & $\mathbf{B{=}5}$ & $\mathbf{B{=}15}$ & $\mathbf{B{=}30}$ & $\mathbf{B{=}60}$ & $\mathbf{B{=}120}$ \\
\midrule
\multicolumn{9}{l}{\textit{Open-weight}} \\
\multirow{5}{*}{\qwenicon~Qwen3-8B} & Prec. & 55.13 & 62.72 & 61.15 & 63.77 & 64.73 & 65.42 & \textbf{65.90} \\
 & Rec. & \textbf{32.81} & 27.24 & 30.18 & 24.81 & 19.70 & 15.02 & 11.36 \\
 & $F_{0.5}$ & 48.53 & 49.76 & \textbf{50.74} & 48.53 & 44.42 & 39.14 & 33.63 \\
 & WED & 6.02 & 4.89 & 4.84 & 3.74 & 2.93 & 2.26 & \textbf{1.57} \\
 & Scribendi & $+0.437$ & $+0.361$ & $+0.407$ & $+0.336$ & $+0.270$ & $+0.201$ & $+0.158$ \\
\midrule
\multicolumn{9}{l}{\textit{High-capacity}} \\
\multirow{5}{*}{\openaiicon~GPT-5.4} & Prec. & 57.58 & 58.96 & 61.42 & 62.98 & 63.38 & 63.96 & \textbf{65.74} \\
 & Rec. & \textbf{48.40} & 47.61 & 45.67 & 42.81 & 40.90 & 39.66 & 38.40 \\
 & $F_{0.5}$ & 55.47 & 56.28 & 57.46 & \textbf{57.55} & 57.10 & 56.98 & 57.55 \\
 & WED & 7.82 & 7.65 & 7.15 & 6.58 & 6.34 & 6.16 & \textbf{5.82} \\
 & Scribendi & $+0.518$ & $+0.497$ & $+0.479$ & $+0.458$ & $+0.440$ & $+0.433$ & $+0.416$ \\
\lightmodelrule
\multirow{5}{*}{\claudeicon~Claude Opus 4.6} & Prec. & 66.50 & 67.98 & 68.79 & 69.80 & 69.07 & \textbf{70.02} & 68.97 \\
 & Rec. & \textbf{41.33} & 40.61 & 40.36 & 39.55 & 37.92 & 38.30 & 38.71 \\
 & $F_{0.5}$ & 59.28 & 59.90 & 60.29 & \textbf{60.54} & 59.32 & 60.07 & 59.64 \\
 & WED & 6.09 & 5.83 & 5.73 & 5.64 & \textbf{5.44} & 5.44 & 5.56 \\
 & Scribendi & $+0.411$ & $+0.398$ & $+0.387$ & $+0.388$ & $+0.379$ & $+0.371$ & $+0.372$ \\
\lightmodelrule
\multirow{5}{*}{\geminiicon~Gemini 3.1-Pro} & Prec. & 58.96 & 60.25 & 60.23 & 62.96 & 63.88 & 64.15 & \textbf{64.89} \\
 & Rec. & \textbf{52.49} & 51.40 & 50.42 & 45.22 & 42.53 & 40.47 & 37.80 \\
 & $F_{0.5}$ & 57.54 & 58.25 & 57.97 & \textbf{58.38} & 58.05 & 57.43 & 56.76 \\
 & WED & 8.85 & 8.59 & 8.56 & 7.34 & 6.72 & 6.50 & \textbf{6.01} \\
 & Scribendi & $+0.544$ & $+0.513$ & $+0.501$ & $+0.434$ & $+0.406$ & $+0.373$ & $+0.347$ \\
\midrule
\multicolumn{9}{l}{\textit{Medium-capacity}} \\
\multirow{5}{*}{\openaiicon~GPT-4.1-mini} & Prec. & 55.08 & 55.74 & 58.16 & 61.02 & 61.81 & 62.88 & \textbf{63.80} \\
 & Rec. & 44.21 & \textbf{44.72} & 42.89 & 41.58 & 40.25 & 39.71 & 37.39 \\
 & $F_{0.5}$ & 52.50 & 53.12 & 54.29 & 55.80 & 55.83 & \textbf{56.31} & 55.90 \\
 & WED & 7.89 & 7.92 & 7.18 & 6.56 & 6.39 & 6.15 & \textbf{5.76} \\
 & Scribendi & $+0.401$ & -- & -- & $+0.365$ & $+0.354$ & $+0.344$ & $+0.331$ \\
\lightmodelrule
\multirow{5}{*}{\claudeicon~Claude Sonnet 4.6} & Prec. & 62.66 & 65.26 & \textbf{66.06} & 65.02 & 65.74 & 65.29 & 64.43 \\
 & Rec. & 36.48 & 36.41 & 37.07 & 39.46 & 39.13 & 40.53 & \textbf{40.70} \\
 & $F_{0.5}$ & 54.79 & 56.34 & 57.12 & 57.56 & 57.87 & \textbf{58.18} & 57.70 \\
 & WED & 5.95 & \textbf{5.70} & 5.72 & 6.22 & 6.06 & 6.28 & 6.46 \\
 & Scribendi & $+0.469$ & -- & -- & $+0.426$ & $+0.391$ & $+0.396$ & $+0.382$ \\
\lightmodelrule
\multirow{5}{*}{\geminiicon~Gemini 3-Flash} & Prec. & 54.67 & 55.28 & 55.46 & 55.79 & \textbf{56.31} & 55.29 & 53.41 \\
 & Rec. & 53.13 & 53.26 & 53.12 & 53.12 & \textbf{53.32} & 53.26 & 53.30 \\
 & $F_{0.5}$ & 54.35 & 54.87 & 54.97 & 55.23 & \textbf{55.68} & 54.87 & 53.39 \\
 & WED & 9.48 & 9.41 & 9.33 & 9.38 & \textbf{9.28} & 9.47 & 9.79 \\
 & Scribendi & $+0.492$ & -- & -- & $+0.460$ & $+0.445$ & $+0.445$ & $+0.436$ \\
\bottomrule
\end{tabular}
\caption{\textbf{Batching sweep on BEA-2019 dev under the best manual prompt, minimal-edits few-shot + taxonomy (Appendix~\ref{prompt:taxonomy-few-shot}).} For each model we report Precision, Recall, ERRANT $F_{0.5}$ (\%), word edit distance (WED, \%), and Scribendi score \citep{islam-magnani-2021-scribendi} (reference-free fluency metric in $[-1,+1]$, computed with Gemma 2 9B as the scoring LM following the MultiGEC-2025 protocol \citep{masciolini2025multigec}; scores each predicted sentence as $+1$ if fluency improved and surface similarity $\geq 0.8$, $0$ if unchanged, $-1$ otherwise; corpus-level mean). A dash (--) marks a Scribendi score not computed for that configuration. Per-model highest Precision, Recall, and $F_{0.5}$ and lowest WED are bolded.}
\label{tab:batching-best-prompt}
\end{table*}

\begin{figure*}[t!]
\centering
\includegraphics[width=\textwidth]{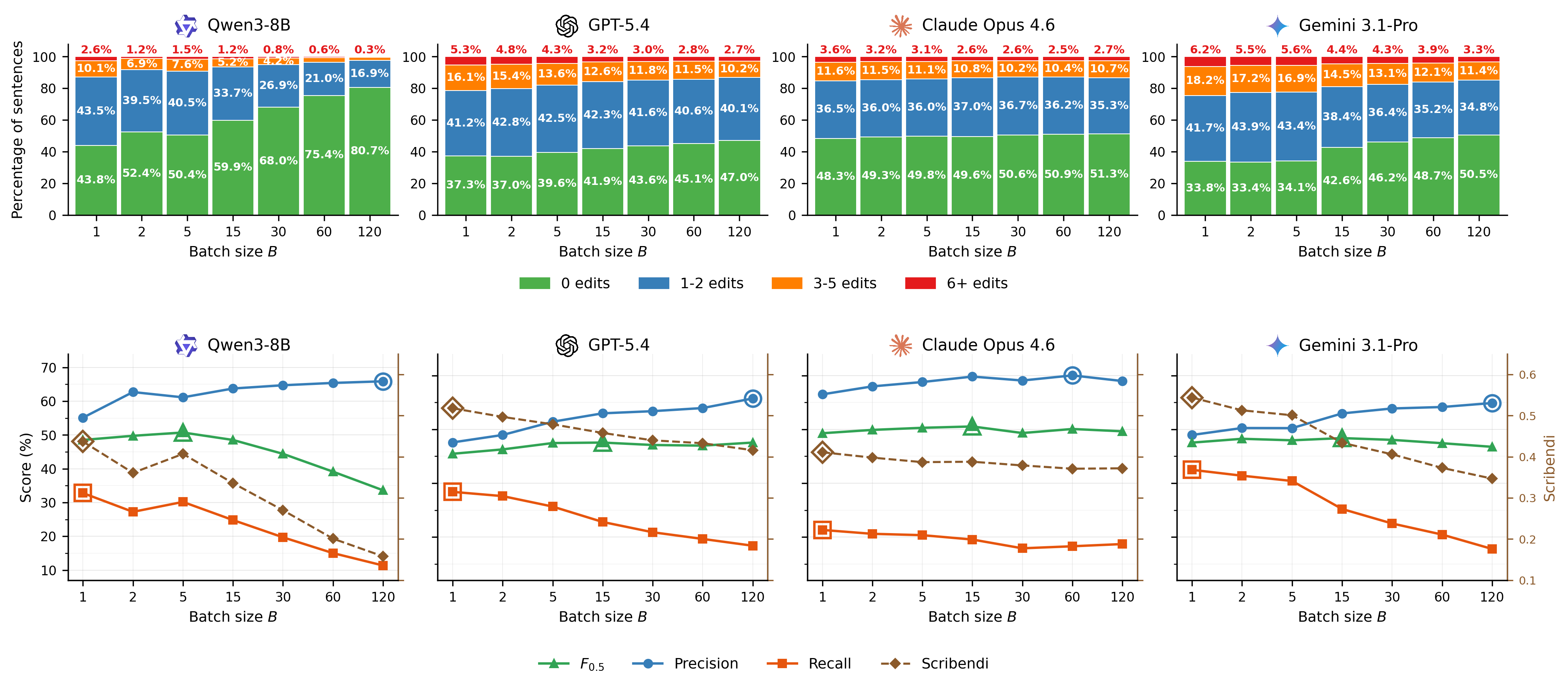}
\caption{\textbf{Per-sentence edit-count distributions and per-model batching-trend
metrics on BEA-2019 dev under our top performing manual prompt, minimal-edits few-shot +
taxonomy (\ref{prompt:taxonomy-few-shot}), for Qwen3-8B and the three high-capacity models
(GPT-5.4, Claude Opus 4.6, Gemini 3.1-Pro), across batch sizes $B \in \{1, 2, 5, 15, 30, 60,
120\}$.} Top row: per-sentence word-level edit counts, bucketed into 0, 1--2, 3--5, and 6+
edits. Bottom row: Precision (blue circles), Recall (orange squares), and $F_{0.5}$ (green
triangles) on the left axis, and the Scribendi score on the right axis (brown dashed
diamonds), with open rings marking the per-model Precision,
Recall, $F_{0.5}$, and Scribendi maxima. This is the same figure as
Figure~\ref{fig:edit-distribution-best-prompt} of Section~\ref{sec:exp-batching}, reproduced
here next to the corresponding numerical results in
Table~\ref{tab:edit-distribution-best-prompt} (edit-count distributions) and
Table~\ref{tab:batching-best-prompt} (batching-trend metrics), for the shared batch sizes
$B \in \{1, 2, 5, 15, 30, 60, 120\}$.}
\label{fig:batching-combined-appendix}
\end{figure*}

\begin{figure*}[t]
\centering
\includegraphics[width=\textwidth]{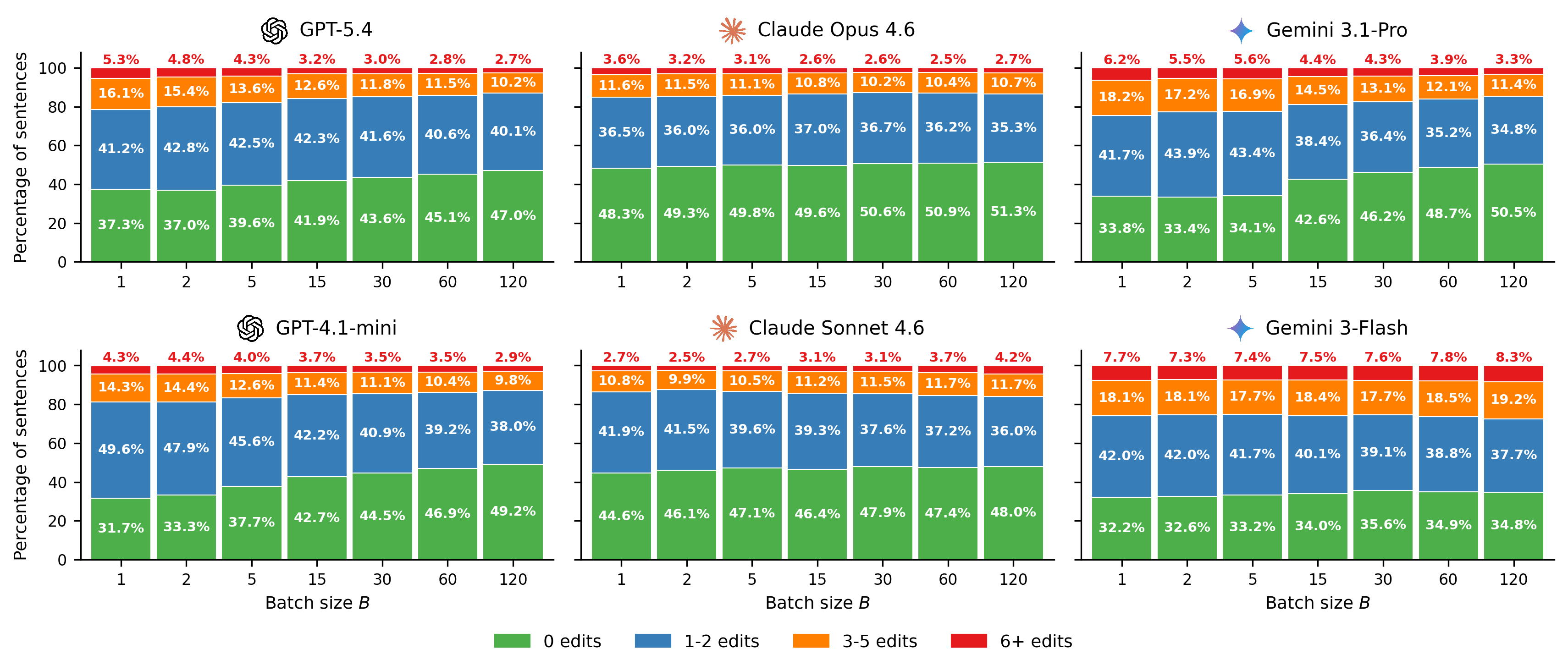}
\caption{\textbf{Per-sentence word-level edit counts across batch sizes on BEA-2019 dev
(4{,}384 sentences) under the best manual prompt, minimal-edits few-shot + taxonomy
(\ref{prompt:taxonomy-few-shot}), all six models.} All models are shown at
$B \in \{1, 2, 5, 15, 30, 60, 120\}$. Edits are word-level substitutions,
deletions, and insertions between the spaCy \texttt{en\_core\_web\_sm}-tokenized source and
the model output, binned per sentence into 0, 1--2, 3--5, and 6+ edits. As the batch size
$B$ grows, the general tendency on five of the six models is to shift mass into the 0-edit
bucket (green), with the largest relative reductions in the 6+ tail (red) and the 3--5
bucket (orange). Claude Sonnet~4.6 is the exception: its 6+ tail rises with $B$, while mass
moves out of the 1--2 bucket into the 0, 3--5, and 6+ buckets, so the model becomes more
decisive at both ends. The high-capacity subset
appears as Figure~\ref{fig:edit-distribution-best-prompt} of
Section~\ref{sec:exp-batching}.}
\label{fig:edit-distribution-best-prompt-6}
\end{figure*}

To trace the mechanism behind the $F_{0.5}$ and WED trends reported above, we also
report the distribution of per-sentence word-level edits introduced by each model at every
batch size on BEA-2019 dev under the best manual prompt, minimal-edits few-shot + taxonomy
(\ref{prompt:taxonomy-few-shot}). For every (model,
batch) cell we tokenize the
source and the model output with spaCy \texttt{en\_core\_web\_sm} and compute
the per-sentence word-level Levenshtein distance ($S{+}D{+}I$) between source
and prediction. Sentences are then bucketed into 0, 1--2, 3--5, and 6+ edits.
The open-weight and high-capacity subset of the resulting distribution is shown as
the top row of Figure~\ref{fig:edit-distribution-best-prompt} of Section~\ref{sec:exp-batching},
whose bottom row reports the corresponding batching-trend metrics; the full six-model version
appears as Figure~\ref{fig:edit-distribution-best-prompt-6}, and
Table~\ref{tab:edit-distribution-best-prompt} lists the underlying percentages,
mean edit count, and maximum edit count for every (model, batch) cell.

\begin{table*}[t]
\centering
\small
\setlength{\tabcolsep}{4pt}
\begin{tabular}{l l rrrrrrr}
\toprule
\textbf{Model} & \textbf{Bucket / stat} & $\mathbf{B{=}1}$ & $\mathbf{B{=}2}$ & $\mathbf{B{=}5}$ & $\mathbf{B{=}15}$ & $\mathbf{B{=}30}$ & $\mathbf{B{=}60}$ & $\mathbf{B{=}120}$ \\
\midrule
\multicolumn{9}{l}{\textit{Open-weight}} \\
\multirow{6}{*}{Qwen3-8B} & 0 edits (\%) & 43.8 & 52.4 & 50.4 & 59.9 & 68.0 & 75.4 & 80.7 \\
 & 1--2 edits (\%) & 43.5 & 39.5 & 40.5 & 33.7 & 26.9 & 21.0 & 16.9 \\
 & 3--5 edits (\%) & 10.1 & 6.9 & 7.6 & 5.2 & 4.2 & 3.0 & 2.1 \\
 & 6+ edits (\%) & 2.6 & 1.2 & 1.5 & 1.2 & 0.8 & 0.6 & 0.3 \\
 & mean edits & 1.19 & 0.87 & 0.94 & 0.75 & 0.58 & 0.43 & 0.31 \\
 & max edits & 44 & 30 & 45 & 58 & 42 & 30 & 29 \\
\midrule
\multicolumn{9}{l}{\textit{High-capacity}} \\
\multirow{6}{*}{GPT-5.4} & 0 edits (\%) & 37.3 & 37.0 & 39.6 & 41.9 & 43.6 & 45.1 & 47.0 \\
 & 1--2 edits (\%) & 41.2 & 42.8 & 42.5 & 42.3 & 41.6 & 40.6 & 40.1 \\
 & 3--5 edits (\%) & 16.1 & 15.4 & 13.6 & 12.6 & 11.8 & 11.5 & 10.2 \\
 & 6+ edits (\%) & 5.3 & 4.8 & 4.3 & 3.2 & 3.0 & 2.8 & 2.7 \\
 & mean edits & 1.64 & 1.58 & 1.46 & 1.33 & 1.26 & 1.22 & 1.15 \\
 & max edits & 48 & 47 & 49 & 44 & 36 & 37 & 35 \\
\midrule[\heavyrulewidth]
\multirow{6}{*}{Claude Opus 4.6} & 0 edits (\%) & 48.3 & 49.3 & 49.8 & 49.6 & 50.6 & 50.9 & 51.3 \\
 & 1--2 edits (\%) & 36.5 & 36.0 & 36.0 & 37.0 & 36.7 & 36.2 & 35.3 \\
 & 3--5 edits (\%) & 11.6 & 11.5 & 11.1 & 10.8 & 10.2 & 10.4 & 10.7 \\
 & 6+ edits (\%) & 3.6 & 3.2 & 3.1 & 2.6 & 2.6 & 2.5 & 2.7 \\
 & mean edits & 1.28 & 1.22 & 1.19 & 1.15 & 1.12 & 1.11 & 1.13 \\
 & max edits & 76 & 46 & 44 & 46 & 44 & 40 & 41 \\
\midrule[\heavyrulewidth]
\multirow{6}{*}{Gemini 3.1-Pro} & 0 edits (\%) & 33.8 & 33.4 & 34.1 & 42.6 & 46.2 & 48.7 & 50.5 \\
 & 1--2 edits (\%) & 41.7 & 43.9 & 43.4 & 38.4 & 36.4 & 35.2 & 34.8 \\
 & 3--5 edits (\%) & 18.2 & 17.2 & 16.9 & 14.5 & 13.1 & 12.1 & 11.4 \\
 & 6+ edits (\%) & 6.2 & 5.5 & 5.6 & 4.4 & 4.3 & 3.9 & 3.3 \\
 & mean edits & 1.82 & 1.74 & 1.71 & 1.46 & 1.35 & 1.27 & 1.18 \\
 & max edits & 41 & 52 & 51 & 44 & 43 & 39 & 36 \\
\midrule
\multicolumn{9}{l}{\textit{Medium-capacity}} \\
\multirow{6}{*}{GPT-4.1-mini} & 0 edits (\%) & 31.7 & 33.3 & 37.7 & 42.7 & 44.5 & 46.9 & 49.2 \\
 & 1--2 edits (\%) & 49.6 & 47.9 & 45.6 & 42.2 & 40.9 & 39.2 & 38.0 \\
 & 3--5 edits (\%) & 14.3 & 14.4 & 12.6 & 11.4 & 11.1 & 10.4 & 9.8 \\
 & 6+ edits (\%) & 4.3 & 4.4 & 4.0 & 3.7 & 3.5 & 3.5 & 2.9 \\
 & mean edits & 1.58 & 1.57 & 1.44 & 1.33 & 1.28 & 1.23 & 1.14 \\
 & max edits & 51 & 46 & 46 & 44 & 41 & 40 & 40 \\
\midrule[\heavyrulewidth]
\multirow{6}{*}{Claude Sonnet 4.6} & 0 edits (\%) & 44.6 & 46.1 & 47.1 & 46.4 & 47.9 & 47.4 & 48.0 \\
 & 1--2 edits (\%) & 41.9 & 41.5 & 39.6 & 39.3 & 37.6 & 37.2 & 36.0 \\
 & 3--5 edits (\%) & 10.8 & 9.9 & 10.5 & 11.2 & 11.5 & 11.7 & 11.7 \\
 & 6+ edits (\%) & 2.7 & 2.5 & 2.7 & 3.1 & 3.1 & 3.7 & 4.2 \\
 & mean edits & 1.19 & 1.13 & 1.15 & 1.24 & 1.21 & 1.27 & 1.30 \\
 & max edits & 46 & 32 & 40 & 49 & 44 & 45 & 47 \\
\midrule[\heavyrulewidth]
\multirow{6}{*}{Gemini 3-Flash} & 0 edits (\%) & 32.2 & 32.6 & 33.2 & 34.0 & 35.6 & 34.9 & 34.8 \\
 & 1--2 edits (\%) & 42.0 & 42.0 & 41.7 & 40.1 & 39.1 & 38.8 & 37.7 \\
 & 3--5 edits (\%) & 18.1 & 18.1 & 17.7 & 18.4 & 17.7 & 18.5 & 19.2 \\
 & 6+ edits (\%) & 7.7 & 7.3 & 7.4 & 7.5 & 7.6 & 7.8 & 8.3 \\
 & mean edits & 1.96 & 1.94 & 1.92 & 1.92 & 1.90 & 1.93 & 2.02 \\
 & max edits & 53 & 51 & 48 & 48 & 50 & 46 & 47 \\
\bottomrule
\end{tabular}
\caption{\textbf{Per-sentence word-edit distributions on BEA-2019 dev} under the best manual
prompt, minimal-edits few-shot + taxonomy (Appendix~\ref{prompt:taxonomy-few-shot}). For each
(model, batch size $B$) cell we tokenize the source and the model output with spaCy
\texttt{en\_core\_web\_sm}, compute the per-sentence word-level Levenshtein distance
($S{+}D{+}I$) between source and prediction, and bin the 4{,}384 sentences into 0, 1--2, 3--5,
and 6+ edit buckets. Mean and max are over the same 4{,}384 sentences.}
\label{tab:edit-distribution-best-prompt}
\end{table*}

\end{document}